%% file: paper.tex
\documentclass[11pt]{article}
\input{macros}

\newcommand{\papertitle}{Momentum Smooths the Path to Gradient Equilibrium}
\newcommand{\authorJVR}{João Vitor Romano}
\newcommand{\authorMZ}{Margaux Zaffran}
\newcommand{\authorRJT}{Ryan J. Tibshirani}

\title{\papertitle}
\author[1]{\authorJVR}
\author[2]{\authorMZ}
\author[1]{\authorRJT}
\affil[1]{\small Department of Statistics, University of California, Berkeley,
  United States}
\affil[2]{\small CELESTE Project Team, Inria, Laboratoire de Mathématiques
  d'Orsay, Université Paris-Saclay, Orsay, France}
\date{}

\hypersetup{
  pdftitle={\papertitle},
  pdfauthor={\authorJVR, \authorMZ, \authorRJT},
  pdflang={en-US},
  pdfdisplaydoctitle=true
}

\begin{document}
\maketitle

\vspace{-0.5cm}
\input{section/0-abstract}
\input{section/1-introduction}
\input{section/2-preliminaries}
\input{section/3-convergence}
\input{section/4-smoothness}
\input{section/5-empirical-results}
\input{section/6-discussion}

\section*{Acknowledgments}

We are grateful to Renaud Gaucher for various helpful discussions, and to Adrien
Taylor for discussion on first-order optimization and acceleration methods. We
would also like to thank the anonymous reviewers whose feedback improved the
paper.

\bibliographystyle{plainnat}
\bibliography{main,refs}

\clearpage
\appendix
\input{section/7-appendix}

\end{document}

%% file: macros.tex
\usepackage[round]{natbib}
\usepackage{authblk}
\usepackage[left=25mm,right=25mm,top=25mm,bottom=25mm]{geometry}
\usepackage[T1]{fontenc}
\usepackage{newtxtext}
\usepackage{amsmath}
\usepackage{amssymb}
\usepackage{parskip}
\usepackage{mathtools}
\usepackage{amsthm}
\usepackage{xcolor}
\usepackage{graphicx}
  \graphicspath{{./figures/}}
\usepackage[skip=3pt]{subcaption}
\usepackage{siunitx}
\usepackage[ruled,noline,noend]{algorithm2e}
  \DontPrintSemicolon
  \SetAlCapHSkip{0pt}
  \SetAlgoSkip{}
  \SetAlgoCaptionSeparator{}
  \SetKwFor{For}{for}{}{}
  \AddToHook{cmd/@algocf@start/after}{\addtolength{\hsize}{1.5em}}
\usepackage{enumitem}
  \setlist{topsep=0pt,leftmargin=2em}
\usepackage{microtype}

\usepackage{zref-clever}
\zcLanguageSetup{english}{
  lastsep  = {,~and\nobreakspace} ,
  tlastsep = {,~and\nobreakspace} ,
}
\zcsetup{cap=true,abbrev=false,nameinlink=false,nudge=true}
\zcRefTypeSetup{assumption}{
  Name-sg = Assumption,
  name-sg = assumption,
  Name-pl = Assumptions,
  name-pl = assumptions,
}
\AddToHook{env/proposition/begin}{\zcsetup{countertype={theorem=proposition}}}
\AddToHook{env/lemma/begin}{\zcsetup{countertype={theorem=lemma}}}
\AddToHook{env/corollary/begin}{\zcsetup{countertype={theorem=corollary}}}
\AddToHook{env/definition/begin}{\zcsetup{countertype={theorem=definition}}}
\AddToHook{env/assumption/begin}{\zcsetup{countertype={theorem=assumption}}}
\AddToHook{env/remark/begin}{\zcsetup{countertype={theorem=remark}}}
\zcsetup{countertype={algocf=algorithm}}
\newcommand{\cref}[1]{\zcref{#1}}

\usepackage{hyperref}
\hypersetup{colorlinks=true, allcolors=blue}


\usepackage{ragged2e}
\RaggedRight
\makeatletter
\renewcommand{\thm@space@setup}{%
  \setlength{\thm@preskip}{\medskipamount}%
  \setlength{\thm@postskip}{\thm@preskip}%
}
\AddToHook{cmd/th@remark/after}{\thm@space@setup}
\makeatother
\theoremstyle{plain}
\newtheorem{theorem}{Theorem}[section]
\newtheorem{proposition}[theorem]{Proposition}
\newtheorem{lemma}[theorem]{Lemma}
\newtheorem{corollary}[theorem]{Corollary}
\newtheorem{assumption}[theorem]{Assumption}
\theoremstyle{definition}
\newtheorem{definition}[theorem]{Definition}
\theoremstyle{remark}
\newtheorem{remark}[theorem]{Remark}

\DeclareMathOperator{\tr}{tr}
\newcommand{\N}{\mathbb{N}}
\newcommand{\R}{\mathbb{R}}

\newcommand{\T}{\intercal}
\DeclarePairedDelimiterXPP{\norm}[1]{{}}{\lVert}{\rVert}{}{#1}
\DeclarePairedDelimiterXPP{\abs}[1]{{}}{\lvert}{\rvert}{}{#1}
\newcommand{\aux}{\xi}
\newcommand{\gd}{\tilde{\theta}}
\definecolor{myblue}{RGB}{0,0,128}
\definecolor{myorange}{RGB}{252,163,17}

%% file: section/0-abstract.tex
\begin{abstract}
Online gradient descent has recently been shown to satisfy a property called
\emph{gradient equilibrium} for a broad class of loss functions. This means that
the average of the gradients of the losses along the sequence converges to zero.
With appropriate choices of losses, we can leverage this property in order to
debias predictions or calibrate prediction sets, among other applications.
Meanwhile, optimizing gradient descent for gradient equilibrium can produce a
highly jagged path of iterates, which may be undesirable in certain problem
settings. In this work, we show that a class of momentum methods can be applied
to attenuate such volatility, while maintaining gradient equilibrium at
essentially the same rate and under essentially the same assumptions as ordinary
gradient descent. We complement this theory by providing experiments on real
data, which clearly show that momentum provides greater smoothness at nearly
equivalent gradient equilibrium error.
\end{abstract}

%% file: section/1-introduction.tex
\section{Introduction}
\label{sec:intro}

Recently, \citet{angelopoulos2025gradient} proposed a novel criterion in online
learning called \emph{gradient equilibrium}. Given sequences of loss functions
\smash{$(\ell_t)_{t \in \N}$} and iterates \smash{$(\theta_t)_{t \in \N}$}, we
say that the iterates satisfy gradient equilibrium (GEQ) provided
\[
\label{eq:geq}
\tag{GEQ}
\frac{1}{T} \sum_{t=1}^T g_t(\theta_t) \to 0, \quad \text{as $T \to \infty$},
\]
where each $g_t$ is a (sub)gradient of $\ell_t$. As shown in
\citet{angelopoulos2025gradient}, this criterion can represent sequential
notions of familiar statistical goals: for example, achieving \eqref{eq:geq}
with squared loss \smash{$\ell_t(\theta) = \frac12(y_t-\theta)^2$} means that
$\theta_t$ is unbiased for $y_t$, where this is interpreted by taking averages
along the sequence. As another example, for quantile loss $\ell_t(\theta) =
\psi_\tau(y_t-\theta)$ at the level $\tau \in [0,1]$, \eqref{eq:geq} means that
$\theta_t$ lies above $y_t$ with frequency $\tau$, again measured along the
sequence. (In this interpretation, we choose the subgradient $1-\tau$ when
$y_t=\theta_t$.) Furthermore, when the response $y_t$ is replaced by a residual
$y_t-f_t(x_t)$ from a black-box prediction model $f_t$ on arbitrary features
$x_t$, achieving \eqref{eq:geq} in these two examples translates to debiasing or
recalibrating the predictions of $f_t$. Applications such as these will be key
points of motivation for our work.

Beyond these, there are numerous other applications of gradient equilibrium; for
example, the framework can also be used to adjust black-box predictors to
achieve groupwise debiasing or groupwise recalibration guarantees, or more
generally multiaccuracy guarantees \citep{angelopoulos2025gradient}; it can also
be used for multilevel quantile recalibration subject to monotonicity
constraints \citep{ding2025calibrated}. In general, online gradient
descent---which we will often simply refer to as gradient descent
(GD)---achieves \eqref{eq:geq} under weak conditions on the losses at hand. This
theory covers all of the aforementioned applications, under minor restrictions
on the underlying data sequence, permitting distribution drift and even
adversarial behavior in the data generating mechanism.

While it offers broad guarantees in terms of achieving gradient equilibrium, one
shortcoming of gradient descent is that, for large step sizes, it can produce
iterate sequences which are jagged and erratic, changing rapidly from one time
step to the next. From the point of view of post-processing the outputs of a
black-box predictor, this may be undesirable in certain problem settings. For
example, in financial trading, an erratic sequence of predicted returns for a
given asset could induce frequent changes in portfolio holdings, which would
then incur large transaction costs. Naturally, to mitigate this issue, a user
could simply decrease the step size, and accordingly reduce the volatility of
gradient descent iterates. However, this slows the convergence to gradient
equilibrium, and presents a clear tradeoff: larger step sizes mean tighter
control on \smash{$\norm{\frac1T \sum_{t=1}^T g_t(\theta_t)}_2$} (the gradient
equilibrium error), but smaller step sizes lead to smoother iterate sequences:
smaller fluctuations along $\theta_t,\theta_{t+1},\theta_{t+2},\ldots$.

\begin{figure}[t]
\centering
\includegraphics{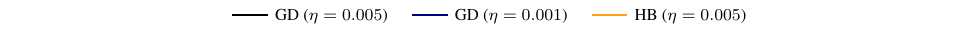}

\begin{subfigure}[b]{0.5\textwidth}
\centering
\includegraphics{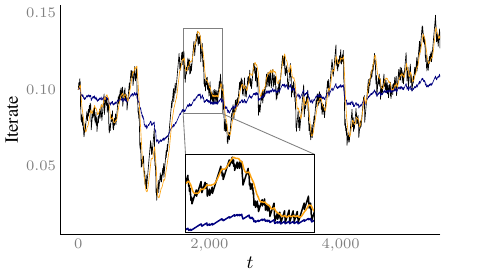}
\caption{Quantile loss: iterates.}
\label{fig:iterates-quantile-loss}
\end{subfigure}%
\begin{subfigure}[b]{0.5\textwidth}
\centering
\includegraphics{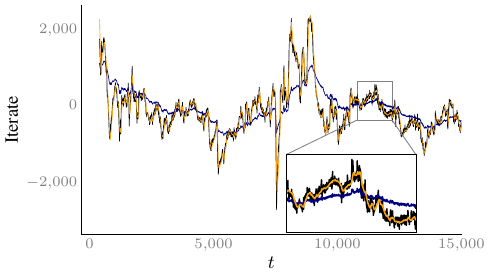}
\caption{Squared loss: iterates.}
\label{fig:iterates-squared-loss}
\end{subfigure}

\medskip
\begin{subfigure}[b]{0.5\textwidth}
\centering
\includegraphics{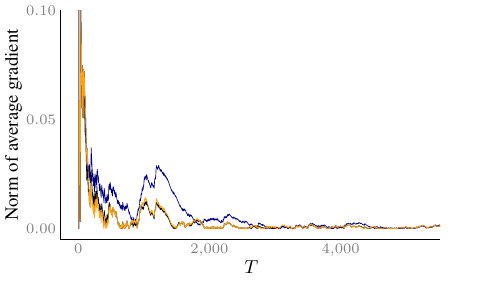}
\caption{Quantile loss: norm of average gradients.}
\label{fig:gradients-quantile-loss}
\end{subfigure}%
\begin{subfigure}[b]{0.5\textwidth}
\centering
\includegraphics{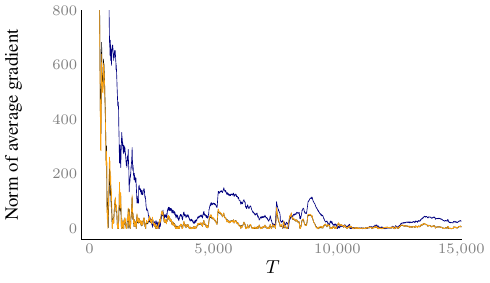}
\caption{Squared loss: norm of average gradients.}
\label{fig:gradients-squared-loss}
\end{subfigure}
\caption{Iterate sequences (top) and gradient equilibrium error (bottom) for
  examples with quantile loss (left) and squared loss (right). Gradient descent
  (GD) with step size $\eta = 0.005$ (black) converges quickly to gradient
  equilibrium, but produces jagged iterate paths. Heavy ball (HB) with the same
  step size and momentum parameter $\beta = 0.95$ (\textcolor{myorange}{orange})
  smooths out these jagged oscillations while leaving the gradient equilibrium
  error intact (the \textcolor{myorange}{orange} and black curves in the
  bottom-row plots largely overlap). Decreasing the step size in gradient
  descent to $\eta = 0.001$ (\textcolor{myblue}{blue}) clearly hinders
  convergence to gradient equilibrium.}
\label{fig:introduction}
\end{figure}

In this paper, we show that a class of momentum methods can be used to
effectively navigate this tradeoff. These methods achieve nearly the same
gradient equilibrium error as ordinary gradient descent, but at a much lower
cost in terms of volatility. \cref{fig:introduction} provides motivating
examples. The left column shows an example with quantile loss, and the right
column an example with squared loss. The top row plots the iterates themselves,
and the bottom row the gradient equilibrium error. In either case, quantile or
squared loss, we can see that gradient descent with step size $\eta=0.005$
produces jagged iterates (black curve), but fast convergence to gradient
equilibrium. Decreasing the step size to $\eta=0.001$ (\textcolor{myblue}{blue}
curve) dampens the iterate dynamics considerably, and hinders the convergence to
gradient equilibrium. Meanwhile, the heavy ball (HB) method with $\eta=0.005$, a
central character in the class of momentum methods we study, attenuates much of
the local oscillation while qualitatively maintaining the iterate dynamics
(\textcolor{myorange}{orange} curve). Importantly, it results in gradient
equilibrium error nearly indistinguishable from that of gradient descent with
the same step size. This figure is meant to be illustrative; our theory and
experiments presented later in the paper confirm that, on these datasets,
\emph{no} choice of step size $\eta$ in gradient descent balances gradient
equilibrium error and iterate smoothness as well as heavy ball.

\paragraph{Contributions summary.}

An outline of our contributions is as follows.

\begin{itemize}
\item We identify a general class of momentum methods (\cref{sec:momentum})
  which encapsulates several standard algorithms such as the heavy ball method.

\item We derive gradient equilibrium guarantees (\cref{sec:geq}) over this class
  in such a way that generalizes existing analyses of gradient descent.

\item We define a criterion for iterate smoothness in terms of discrete second
  derivatives, and prove that momentum generally leads to smoother iterate paths
  than gradient descent (\cref{sec:smoothness}).

\item We conduct experiments on real data to support our theory, which clearly
  show that momentum preserves gradient equilibrium while smoothing iterate
  paths and maintaining the original iterate dynamics (\cref{sec:exp}).
\end{itemize}

Together, our results demonstrate a key benefit of momentum: it strictly
improves the GEQ--smoothness Pareto frontier over ordinary gradient descent,
both for the upper bounds analyzed in \cref{sec:pareto-hb} (see
\cref{fig:pareto-frontier}) and in our experiments on real-world data (see
\cref{fig:frontier-quantile-loss, fig:frontier-squared-loss}).

\paragraph{Related work.}

Momentum methods have deep roots in traditional (batch) optimization, dating
back to the heavy ball method proposed by \citet{polyak1964methods}, and
Nesterov acceleration by \citet{nesterov1983method}. The goal of this line of
work is to achieve faster convergence rates, which Nesterov acceleration
accomplishes definitively: it achieves the optimal convergence rate among all
first-order methods for smooth convex optimization
\citep{nesterov2004introductory}. While momentum methods have been adapted to
stochastic optimization, with of course a major success being Adam
\citep{kingma2015adam}, they are not typically used to induce greater smoothness
in iterate trajectories. In fact, we are unaware of any work that studies this
precisely in stochastic or online settings.

At first glance, the use of momentum schemes such as heavy ball or Nesterov in
order to stabilize iterate paths in online optimization may seem surprising,
because in the batch setting, momentum is actually known to lead to oscillatory
behavior (sometimes called ``Nesterov ripples'' in the context of Nesterov
acceleration). We believe that the success of the class of methods we study in
smoothing the iterate trajectory while preserving its global dynamics relies
crucially on the particular moving average parametrization that we adopt, as in
\citet{sebbouh2021almost}. In fact, this particular moving average
parametrization was also used in the further-away literature of distributed
optimization against faulty or adversarial workers. Notably,
\citet{karimireddy2021learning} and \citet{farhadkhani2022byzantine} used the
same type of parameters within heavy ball in order to reduce noise in workers'
updates without degrading performance too much.

Closely related to our goal is the study of online learning under
\emph{switching costs} \citep{chen2018smoothed, zhang2021revisiting}, where the
learner incurs both a ``hitting cost'', due to the loss of the current action,
and a ``switching cost'', due to the change in action from the previous round.
This exhibits a tradeoff analogous to that in our paper: aggressive updates can
drive down the hitting cost but inflate the switching cost, which may hurt
downstream applications. More broadly related to our goal is the idea of
``stability after improvement'' in operational machine learning systems. This is
a key motivation for the use of \emph{churn constraints}, which control the
deviation of a predictor from a reference model \citep{milanifard2016launch,
goh2016satisfying}. Such constraints can be used to stabilize a model in
deployment that is being used to make decisions, and is updated regularly as new
data becomes available.

Finally, the work on gradient equilibrium by \citet{angelopoulos2025gradient}
grew out of the online conformal prediction literature, initiated by
\citet{gibbs2021adaptive}, with numerous contributions that followed, such as
\citet{zaffran2022adaptive, angelopoulos2023conformal, bhatnagar2023improved,
feldman2023achieving, gibbs2024conformal, podkopaev2024adaptive}. Indeed, a
similar approach to the momentum updates we study in this paper was explored in
the early stages of this literature by \citet{gibbs2021adaptive}, but without
any formal analysis. In a sense, our work provides theoretical backing to close
the loop.

%% file: section/2-preliminaries.tex
\section{Generalized momentum methods}
\label{sec:momentum}

Given a sequence of loss functions \smash{$(\ell_t)_{t \in \N}$}, with each
$\ell_t \colon \R^d \to \R$, we consider a general class of first-order methods
which, given an initialization $\theta_1 \in \R^d$, perform updates of the form:
\[
\label{eq:gmm}
\tag{GMM}
\theta_{t+1} = \theta_t - \sum_{k=1}^t \gamma_{tk} g_k(\theta_k),
\]
for step sizes $\gamma_{tk}$, $1 \leq k \leq t$. We call this a
\emph{generalized momentum method} (see \citealp{drori2014performance,
kim2016optimized} for an analog in the batch setting).

Recall that, as in the introduction, each $g_k(\theta_k)$ in \eqref{eq:gmm}
denotes a (sub)gradient of $\ell_k$ at $\theta_k$. Therefore, if $\gamma_{tk} =
\eta_t$ for $t=k$ and $\gamma_{tk} = 0$ otherwise, then \eqref{eq:gmm} recovers
the standard online gradient descent update with step size $\eta_t$. This
framework can also represent other canonical methods, as we describe below.

\paragraph{Heavy ball.}

For $\gamma_{tk} = \eta_k (1-\beta) \beta^{t-k}$, where $\beta \in [0,1)$ is a
momentum parameter, note that \eqref{eq:gmm} becomes
\begin{equation}
\label{eq:hb}
\theta_{t+1} = \theta_t + \beta (\theta_t-\theta_{t-1}) - \eta_t (1-\beta)
g_t(\theta_t),
\end{equation}
where we set $\theta_0 = \theta_1$ for notational simplicity. This is a version
of the heavy ball method, as studied in \citet{sebbouh2021almost}.
Interestingly, when each $\eta_t = \eta$, this is equivalent to
\begin{align*}
m_t &= \beta m_{t-1} + (1-\beta) g_t(\theta_t), \\
\theta_{t+1} &= \theta_t - \eta m_t,
\end{align*}
where $m_0 = 0$. We can see that $\beta$ weights past gradients (encapsulated in
the momentum term $m_{t-1}$) against the current gradient.

\paragraph{Nesterov acceleration.}

For $\gamma_{tk} = \eta_k (1-\beta) (\mathbb{I}\{t = k\} + \beta^{t-k+1})$,
where $\mathbb{I}\{t = k\} = 1$ for $t=k$ and $\mathbb{I}\{t = k\} = 0$
otherwise, and $\beta \in [0,1)$ is again a momentum parameter, note that
\eqref{eq:gmm} represents the auxiliary sequence associated with the updates
\[
\theta_{t+1} = \theta_t + \beta (\theta_t - \theta_{t-1}) - \eta_t (1-\beta)
g_t(\theta_t + \beta (\theta_t - \theta_{t-1})),
\]
where again $\theta_0 = \theta_1$. This is a version of Nesterov acceleration,
and it can be written equivalently as
\begin{equation}
\label{eq:nesterov}
\begin{aligned}
\omega_t &= \theta_t + \beta (\theta_t - \theta_{t-1}), \\
\theta_{t+1} &= \omega_t - \eta_t (1-\beta) g_t(\omega_t),
\end{aligned}
\end{equation}
which mirrors the traditional formulation of Nesterov acceleration in the batch
setting. In what follows, we focus on the auxiliary sequence
\smash{$(\omega_t)_{t \in \N}$}, which we will generally refer to as
\emph{Nesterov momentum}. As it turns out, this auxiliary sequence is more
amenable to our analysis of gradient equilibrium.

\paragraph{Comparison to traditional batch definitions.}

It is worth noting that our definitions of heavy ball and Nesterov acceleration
above are slightly different from the most direct translations of their
traditional batch definitions to the online setting. In the batch heavy ball
method, the momentum parameter $\beta$ traditionally appears in the update for
$\theta_{t+1}$ as a factor multiplying the iterate difference
$\theta_t-\theta_{t-1}$. In our formulation \eqref{eq:hb}, we also multiply the
gradient $g_t(\theta_t)$ by $1-\beta$. Similarly, the same multiplicative factor
appears in Nesterov momentum. While this difference may seem minor, it is
important: adopting the parametrizations defined above is key to our results,
particularly in terms of providing clearer benefits with respect to smoothness,
both practically and theoretically. Besides, we note that in batch Nesterov
acceleration, one traditionally uses a carefully chosen sequence of time-varying
momentum parameters converging to 1 as $t \to \infty$ (leading to optimal
convergence rates for smooth convex functions). In our formulations
\eqref{eq:hb} and \eqref{eq:nesterov}, though, we use a constant parameter
$\beta \in [0,1)$, as our focus is \emph{not} on assuring optimal convergence
rates toward a minimizer.

%% file: section/3-convergence.tex
\section{Convergence to gradient equilibrium}
\label{sec:geq}

We show that \eqref{eq:gmm} achieves \eqref{eq:geq} for restorative, Lipschitz
losses, which include quantile and logistic losses. This analysis also requires
certain conditions on the step sizes, which are satisfied by heavy ball and
Nesterov momentum. For squared loss, which is non-Lipschitz, we give a more
specialized analysis to establish that heavy ball achieves \eqref{eq:geq}.

\subsection{Preliminaries}

Our analysis of gradient equilibrium is made possible by controlling the total
weight that \eqref{eq:gmm} assigns to past gradients, as the next assumption on
the step sizes formalizes.

\begin{assumption}
\label{ass:bound-stepsize-scheme}
Assume that $\gamma_{tk} \geq 0$ for all $t,k \in \N$ with $k < t$. Assume also
that \smash{$\sum_{k=1}^{t-1} \sum_{u=t}^\infty \gamma_{uk} \leq \delta$}
for a constant $\delta \geq 0$ and all $t \in \N$.
\end{assumption}

To visualize \cref{ass:bound-stepsize-scheme}, consider that $\delta$ must bound
the following infinite block sums depicted in \textcolor{blue}{blue ($t=2$)} and
\textcolor{myorange}{orange ($t=3$)}, as well as all other analogous sums for
the remaining $t=4,5,6,\ldots$:
\[
\begin{bmatrix}
\gamma_{11}                   & 0           & 0           & \ldots \\
\textcolor{blue}{\gamma_{21}} & \gamma_{22} & 0           & \ldots \\
\textcolor{blue}{\gamma_{31}} & \gamma_{32} & \gamma_{33} & \ldots \\
\textcolor{blue}{\gamma_{41}} & \gamma_{42} & \gamma_{43} & \ldots \\
\textcolor{blue}{\vdots}      & \vdots      & \vdots      & \ddots
\end{bmatrix}
\quad \text{and} \quad
\begin{bmatrix}
\gamma_{11}                       & 0                                 & 0           & \ldots \\
\gamma_{21}                       & \gamma_{22}                       & 0           & \ldots \\
\textcolor{myorange}{\gamma_{31}} & \textcolor{myorange}{\gamma_{32}} & \gamma_{33} & \ldots \\
\textcolor{myorange}{\gamma_{41}} & \textcolor{myorange}{\gamma_{42}} & \gamma_{43} & \ldots \\
\textcolor{myorange}{\vdots}      & \textcolor{myorange}{\vdots}      & \vdots      & \ddots
\end{bmatrix}.
\]

Both heavy ball and Nesterov momentum satisfy \cref{ass:bound-stepsize-scheme}
(see \cref{sec:hb} and \cref{app:nesterov}, respectively). Since gradient
descent has $\gamma_{tk} = 0$ whenever $t \neq k$, it also trivially satisfies
\cref{ass:bound-stepsize-scheme} with $\delta=0$.

\begin{definition}
For each $k \in \N$, let $\rho_k = \sum_{u=k}^\infty \gamma_{uk}$ denote the
\emph{total weight} attributed to $g_k(\theta_k)$ by the update at time $k$ and
all later updates.
\end{definition}

Under \cref{ass:bound-stepsize-scheme}, for any $k \in \N$, we have that
$\rho_k$ is finite and satisfies
$\gamma_{kk} \leq \rho_k \leq \gamma_{kk} + \delta$.

To bound the average gradient, we now introduce our second assumption.

\begin{assumption}
\label{ass:positive-total-weights}
Assume that $\rho_t > 0$ for all $t \in \N$.
\end{assumption}

\begin{remark}
For gradient descent, $g_t(\theta_t)$ receives weight $\eta_t$ at time $t$ and
zero weight in all later updates, so $\rho_t = \eta_t$. Importantly, for heavy
ball and Nesterov momentum, later updates may also assign weight to this
gradient, but its total weight is still $\rho_t = \eta_t$ and is independent of
$\beta$ (see \cref{lem:hb,lem:nesterov}). Thus, for those three methods of
interest, \cref{ass:positive-total-weights} is equivalent to $\eta_t > 0$ for
all $t \in \N$. More generally, \cref{ass:positive-total-weights} holds under
\cref{ass:bound-stepsize-scheme} if each update assigns positive weight to its
current gradient, that is, if $\gamma_{tt} > 0$ for all $t \in \N$.
\end{remark}

We can then define an auxiliary sequence \smash{$(\aux_t)_{t \in \N}$} by
$\aux_1 = \theta_1$, and updates of the form:
\begin{equation}
\label{eq:aux}
\aux_{t+1} = \aux_t - \rho_t g_t(\theta_t).
\end{equation}

This update takes a gradient descent step from $\aux_t$ using the gradient of
$\ell_t$ at the \eqref{eq:gmm} iterate $\theta_t$. This auxiliary sequence
depends only on what happens at time $t$, making it easier to bound its growth
than working directly on the sequence $\theta_t$ which keeps track of all
previous gradients. Moreover, by construction, it directly allows us to bound
the gradient equilibrium error as the next result shows, thanks to
\cref{ass:bound-stepsize-scheme}. The next result is proved in the appendix, as
is the default for all results in this paper.

\begin{lemma}
\label{lem:general-redux}
Consider \eqref{eq:gmm}, with arbitrary initialization $\theta_1 \in \R^d$.
Under \cref{ass:positive-total-weights}, for any $T \in \N$,
\begin{align*}
\norm[\bigg]{\frac{1}{T} \sum_{t=1}^T g_t(\theta_t)}_2
&\leq \frac{\norm{\theta_1 - \aux_{T+1}}_2}{\rho_1 T} + \frac{2}{T}
\Bigl(\max_{t \leq T+1} \norm{\aux_t}_2 \Bigr)
\sum_{t=2}^T \abs{\rho_t^{-1} - \rho_{t-1}^{-1}}\\
&\leq \frac{2}{T} \Bigl(\max_{t \leq T+1} \norm{\aux_t}_2 \Bigr)
\biggl(\rho_1^{-1} + \sum_{t=2}^T \abs{\rho_t^{-1} - \rho_{t-1}^{-1}} \biggr),
\end{align*}
where, recall, the auxiliary sequence is as defined in \eqref{eq:aux}.
\end{lemma}

According to \cref{lem:general-redux}, for \eqref{eq:gmm} with a nonincreasing
sequence $(\rho_t)_{t \in \N}$, the gradient equilibrium error is $O((\max_{t
\leq T+1} \norm{\aux_t}_2)(T\rho_T)^{-1})$. Thus, it remains to control $\max_{t
\leq T+1} \norm{\aux_t}_2$, which we do next.

\subsection{Main convergence results}

In this subsection, we assume that each $\ell_t$ is $L$-Lipschitz, which means
that $\norm{g_t(\theta)}_2 \leq L$ for all $\theta \in \R^d$.

To control the growth of the auxiliary iterates, we use conditions on the
curvature of the negative gradient field, as in
\citet{angelopoulos2025gradient}. In particular, following this work, we say
that a loss function $\ell\colon \R^d \to \R$ is \emph{restorative with weak
curvature} provided that there exists a radius $h\geq0$ such that
\[
g(\theta)^\T \theta \geq 0, \quad \text{for all $\norm{\theta}_2 > h$,
  and all (sub)gradients $g(\theta)$ of $\ell$ at $\theta$}.
\]
Intuitively, this means that outside of a ball of radius $h$, the negative
gradient field points back toward the origin. As in the analysis of gradient
descent, the univariate ($d=1$) and multivariate ($d \geq 2$) cases turn out to
be fundamentally different. Our presentation below is divided accordingly.

\subsubsection{Univariate losses}

In the univariate case, the auxiliary iterates satisfy the following bound for
restorative losses with weak curvature. Here and henceforth, we note that there
is no loss of generality in assuming that the radius $h_t$ corresponding to the
restorativity condition for $\ell_t$ is nondecreasing in $t$. (If not, we could
simply redefine the radius at $t$ by $\max_{s\leq t} h_s$.)

\begin{proposition}
\label{prop:weak-curvature-1d}
Let $d = 1$. Consider \eqref{eq:gmm}, with arbitrary initialization $\theta_1
\in \R$. Under \cref{ass:positive-total-weights, ass:bound-stepsize-scheme},
assume that each $\ell_t$ is $L$-Lipschitz, and restorative with weak curvature,
with radius $h_t$ nondecreasing in $t$. Then, the auxiliary iterates satisfy,
for any $T \in \N$,
\[
\max_{t \leq T+1} \abs{\aux_t}
  \leq \max \{\abs{\theta_1}, h_T + \delta L\} + L \max_{t \leq T} \rho_t.
\]
Therefore, combining with \cref{lem:general-redux}, we have
\[
\abs[\bigg]{\frac{1}{T} \sum_{t=1}^T g_t(\theta_t)}
\leq \frac{\abs{\theta_1}}{\rho_1 T} + \biggl(\frac{1}{\rho_1 T} + \frac{2}{T}
\sum_{t=2}^T \abs{\rho_t^{-1} - \rho_{t-1}^{-1}} \biggr) \Bigl(\max
\{\abs{\theta_1}, h_T + \delta L\} + L \max_{t \leq T} \rho_t \Bigr).
\]
\end{proposition}

If $\rho_t = \rho > 0$ and $h_t = h \geq 0$ for all $t \in \N$, then the average
gradient bound in \cref{prop:weak-curvature-1d} is $O(1/T)$.

Quantile and logistic losses are canonical examples of restorative losses in one
dimension (with mild conditions on the data), as the next result makes precise.
Here and throughout, we define
\[
\psi_\tau(u) = \tau \abs{u} \mathbb{I}\{u \geq 0\} + (1-\tau) \abs{u}
\mathbb{I}\{u < 0\}.
\]

\begin{corollary}
\label{cor:quantile-logistic}
The following losses are $L$-Lipschitz and restorative with weak curvature with
radius $h$, for constants $L,h>0$ depending only on $a,b,\tau,\varepsilon$.
\begin{itemize}
\item Quantile loss $\ell_t(\theta) = \psi_\tau(y_t - \theta)$, where $\abs{y_t}
  \leq b$, and $\tau\in[0,1]$.
\item Generalized logistic loss: $\ell_t(\theta) = -y_t\theta +
  (b-a)\log(1+e^\theta)+a\theta$, where $y_t \in [a+\varepsilon,
  b-\varepsilon]$, and $\varepsilon>0$.
\end{itemize}
In either case, \cref{prop:weak-curvature-1d} applies, and if we further assume
that $(\rho_t)_{t \in \N}$ is nonincreasing, the proposition yields an average
gradient bound
\smash{$\abs{\frac{1}{T} \sum_{t=1}^T g_t(\theta_t)} = O((\rho_T T)^{-1})$}, and
therefore also $O((\gamma_{TT} T)^{-1})$ whenever $\gamma_{TT} > 0$. The latter
is simply $O(1/T)$ for constant step sizes $\gamma_{tt} = \eta > 0$ for all
$t \in \N$.
\end{corollary}

As in \citet{angelopoulos2025gradient}, a notable application of
\cref{cor:quantile-logistic} involves performing online updates using quantile
or generalized logistic loss to adjust a black-box prediction $f_t(x_t)$ of a
response $z_t$. In particular, at each time $t$, we adjust the next prediction
via $f_{t+1}(x_{t+1})+\theta_{t+1}$ in the quantile case, with $\theta_{t+1}$
learned by \eqref{eq:gmm}. In the language of the corollary, this translates to
replacing each instance of $y_t$ by the residual $z_t-f_t(x_t)$. To examine the
assumptions in this setting, for quantile loss, the corollary requires bounded
errors: $\abs{z_t-f_t(x_t)} \leq b$. Meanwhile, for logistic loss, it requires
$z_t-f_t(x_t) \in [a+\varepsilon, b-\varepsilon]$, which holds for binary $z_t$
by assuming $f_t(x_t) \in [\varepsilon, 1-\varepsilon]$ and setting $a=-1$ and
$b=1$. Further, a short calculation reveals that achieving gradient equilibrium
\eqref{eq:geq} here means calibrating the predictor to obtain nominal coverage,
or debiasing the predictor:
\[
\frac{1}{T} \sum_{t=1}^T \mathbb{I}\{z_t \leq f_t(x_t) + \theta_t\} \to \tau,
\quad \text{or} \quad
\frac{1}{T} \sum_{t=1}^T \biggl(f_t(x_t) + a +
(b-a)\frac{e^{\theta_t}}{1+e^{\theta_t}}\biggr) - \frac{1}{T} \sum_{t=1}^T z_t
\to 0,
\]
for quantile loss and logistic loss, respectively. Applications of this form are
considered in \cref{sec:exp}.

\subsubsection{Multivariate losses}

In multiple dimensions, under weak curvature, the auxiliary iterates are no
longer bounded in general, but we can control the rate of their growth.

\begin{proposition}
\label{prop:weak-curvature}
Let $d \geq 2$. Consider \eqref{eq:gmm}, with arbitrary initialization $\theta_1
\in \R^d$. Under \cref{ass:positive-total-weights, ass:bound-stepsize-scheme},
assume that each $\ell_t$ is $L$-Lipschitz and restorative with weak curvature,
with radius $h_t$. Then, the auxiliary iterates satisfy, for any $T \in \N$,
\[
\max_{t \leq T+1} \norm{\aux_t}_2
\leq \sqrt{\norm{\theta_1}_2^2 + L^2 \sum_{t=1}^T \rho_t^2 + 2L \sum_{t=1}^T
  \rho_t (h_t + \delta L)}.
\]
Therefore, combining with \cref{lem:general-redux}, we have
\[
\norm[\bigg]{\frac{1}{T} \sum_{t=1}^T g_t(\theta_t)}_2
\leq \frac{\norm{\theta_1}_2}{\rho_1 T} + \biggl(\frac{1}{\rho_1 T} +
\frac{2}{T} \sum_{t=2}^T \abs{\rho_t^{-1} - \rho_{t-1}^{-1}} \biggr)
\sqrt{\norm{\theta_1}_2^2 + L^2 \sum_{t=1}^T \rho_t^2 + 2L \sum_{t=1}^T \rho_t
  (h_t + \delta L)}.
\]
\end{proposition}

If $\rho_t = \rho > 0$ and $h_t = h$ for all $t \in \N$, then the average
gradient bound in \cref{prop:weak-curvature} is \smash{$O(1/\sqrt{T})$}. For
$d \geq 2$, it can be shown that under the given assumptions, this rate is
unimprovable; this is established in \citet{angelopoulos2025gradient} for
gradient descent, which implies the same is true for generalized momentum
methods (since the former is a special case of the latter).

To obtain the fast $O(1/T)$ rate for gradient equilibrium error in multivariate
problems, we must assume stronger curvature, as we do next. We say that
$\ell\colon \R^d \to \R$ is \emph{restorative with positive curvature} provided
that there exists a radius $h\geq0$ and curvature constant $\phi>0$ such that
\[
g(\theta)^\T \theta \geq \phi, \quad \text{for all $\norm{\theta}_2 > h$,
  and all (sub)gradients $g(\theta)$ of $\ell$ at $\theta$}.
\]
In what follows, we will assume that the curvature $\phi_t$ for $\ell_t$
satisfies $\phi_t \geq (\rho_t/2 + \delta) L^2$. We note that this changes the
nature of the restorativity assumption, compared to the weak case: now, the
assumption is not only on the loss $\ell_t$ but also on the step sizes used in
the update scheme, because $\rho_t$ and $\delta$ (which depend on individual
values of $\gamma_{tk}$) must obey an upper bound in terms of $\phi_t$.

\begin{proposition}
\label{prop:positive-curvature}
Let $d \geq 2$. Consider \eqref{eq:gmm}, with arbitrary initialization $\theta_1
\in \R^d$. Under \cref{ass:positive-total-weights, ass:bound-stepsize-scheme},
assume that each $\ell_t$ is $L$-Lipschitz, and restorative with positive
curvature, with radius $h_t$ nondecreasing in $t$, and curvature satisfying
$\phi_t \geq (\rho_t/2 + \delta) L^2$. Then, the auxiliary iterates satisfy, for
any $T \in \N$,
\[
\max_{t \leq T+1} \norm{\aux_t}_2
  \leq \max \{\norm{\theta_1}_2, h_T + \delta L\} + L \max_{t \leq T} \rho_t.
\]
Therefore, combining with \cref{lem:general-redux}, we have
\[
\norm[\bigg]{\frac{1}{T} \sum_{t=1}^T g_t(\theta_t)}_2
  \leq \frac{\norm{\theta_1}_2}{\rho_1 T} + \biggl(\frac{1}{\rho_1 T} + \frac{2}{T} \sum_{t=2}^T \abs{\rho_t^{-1} - \rho_{t-1}^{-1}} \biggr) \Bigl(\max \{\norm{\theta_1}_2, h_T + \delta L\} + L \max_{t \leq T} \rho_t \Bigr).
\]
\end{proposition}

If $\rho_t = \rho > 0$ and $h_t = h \geq 0$ for all $t \in \N$, then the average
gradient bound in \cref{prop:positive-curvature} is $O(1/T)$. Furthermore, the
forms of the auxiliary iterate and average gradient bounds in
\cref{prop:positive-curvature} are direct multivariate analogs of those given
earlier in \cref{prop:weak-curvature-1d} for the univariate case.

\begin{remark}
For gradient descent with arbitrary positive step sizes, $\rho_t = \eta_t$,
$\aux_t = \theta_t$, and $\delta = 0$. In this case,
\cref{prop:weak-curvature-1d, prop:weak-curvature, prop:positive-curvature}
recover the results in parts (a), (b), and (c) of Proposition 25 in
\citet{angelopoulos2025gradient}, and in this sense, our bounds are a strict
generalization of known results for gradient descent on Lipschitz losses.

However, an important case not covered by our theory above is the squared loss
\smash{$\ell_t(\theta) = \frac12(y_t-\theta)^2$}, which is not Lipschitz.
\citet{angelopoulos2025gradient} provide theory for gradient descent on squared
loss as a special case of a more general theory on \emph{locally} Lipschitz
losses which possess an even stronger form of curvature, called quadratic
curvature. Unfortunately, this local Lipschitz theory does not extend cleanly to
the case of momentum methods. Furthermore, experiments in \cref{app:swing}
suggest that quadratic curvature is not sufficient for controlling gradient
equilibrium error for momentum methods on locally Lipschitz losses in general;
the accumulation of (potentially unbounded) gradients in this setting can cause
iterates to diverge rapidly away from the initialization.
\end{remark}

\subsection{Heavy ball}
\label{sec:hb}

In this subsection, we specialize the previous results to the heavy ball method
as defined in \eqref{eq:hb}. Recall that this is a special case of
\eqref{eq:gmm} with $\gamma_{tk} = \eta_k (1-\beta) \beta^{t-k}$. First we
verify that this method satisfies the assumptions required for the previous
theory, and make the relevant constants explicit.

\begin{lemma}
\label{lem:hb}
Heavy ball \eqref{eq:hb} with nonincreasing positive step sizes $\eta_1 \geq
\eta_2 \geq \cdots$ and $\beta \in [0,1)$ satisfies
\cref{ass:positive-total-weights, ass:bound-stepsize-scheme} with $\delta =
\eta_1 \beta/(1-\beta)$ and $\rho_t = \eta_t$ for all $t \in \N$.
\end{lemma}

With these assumptions in place, we can now translate the previous propositions
over to the heavy ball case.

\begin{corollary}
\label{cor:hb}
Heavy ball \eqref{eq:hb} with nonincreasing positive step sizes $\eta_1 \geq
\eta_2 \geq \cdots$, $\beta \in [0,1)$, and arbitrary initialization $\theta_1
\in \R^d$ satisfies the following average gradient bounds for any $T \in \N$. In
each case, we assume that the losses are $L$-Lipschitz.

\begin{enumerate}
\item[(i)] If $d=1$ and each $\ell_t$ is restorative with weak curvature, with
  radius $h_t$ nondecreasing, then by \cref{prop:weak-curvature-1d},
\[
\abs[\bigg]{\frac{1}{T} \sum_{t=1}^T g_t(\theta_t)}
  \leq \frac{\abs{\theta_1}}{\eta_1 T} + \biggl(\frac{2}{\eta_T T} -
  \frac{1}{\eta_1 T}\biggr) \biggl(\max \Bigl\{\abs{\theta_1}, h_T +
  \frac{\eta_1\beta}{1-\beta} L\Bigr\} + \eta_1 L \biggr).
\]

\item[(ii)] If $d \geq 2$ and each $\ell_t$ is restorative with weak curvature,
  then by \cref{prop:weak-curvature},
\[
\norm[\bigg]{\frac{1}{T} \sum_{t=1}^T g_t(\theta_t)}_2
  \leq \frac{\norm{\theta_1}_2}{\eta_1 T} + \biggl(\frac{2}{\eta_T T} -
  \frac{1}{\eta_1 T}\biggr) \sqrt{\norm{\theta_1}_2^2 + L^2 \sum_{t=1}^T
    \eta_t^2 + 2L \sum_{t=1}^T \eta_t \biggl(h_t + \frac{\eta_1\beta}{1-\beta}
    L\biggr)}.
\]

\item[(iii)] If $d \geq 2$ and each $\ell_t$ is restorative with positive
  curvature, with radius $h_t$ nondecreasing, and curvature satisfying
  $\phi_t \geq (\rho_t/2 + \delta) L^2$, then by \cref{prop:positive-curvature},
\[
\norm[\bigg]{\frac{1}{T} \sum_{t=1}^T g_t(\theta_t)}_2
  \leq \frac{\norm{\theta_1}_2}{\eta_1 T} + \biggl(\frac{2}{\eta_T T} -
  \frac{1}{\eta_1 T}\biggr) \biggl(\max \Bigl\{\norm{\theta_1}_2, h_T +
  \frac{\eta_1\beta}{1-\beta} L\Bigr\} + \eta_1 L \biggr).
\]
\end{enumerate}
\end{corollary}

As in the discussions above (pertaining to the original propositions), in the
case of constant step sizes and radii, $\eta_t = \eta > 0$ and $h_t = h$ for all
$t \in \N$, the average gradient bound in parts (i) and (iii) of \cref{cor:hb}
is $O(1/T)$, and in part (ii) it is \smash{$O(1/\sqrt{T})$}.

\begin{remark}[Choosing $\beta$ for GEQ]
If we focus exclusively on convergence to gradient equilibrium, without
considering iterate smoothness or other metrics such as regret, \cref{cor:hb}
suggests that momentum does not significantly harm gradient equilibrium error
in relevant settings. Indeed, consider constant step sizes $\eta_t = \eta = c
b/L$, where $0 < c \ll 1$ and $b$ is typically a bound on the targets such that
$h_T \approx b$ (one can think, for example, of the quantile loss case). Then,
the heavy ball and gradient descent bounds nearly coincide as long as $h_T/\eta
\approx L/c$ dominates $L/(1 - \beta)$, that is, $1 - \beta \gg c$. For a
choice of $\beta$ that also accounts for the smoothing benefits that momentum
brings, see \cref{sec:pareto-hb}, where we show how momentum improves the
tradeoff between gradient equilibrium error and iterate smoothness.
\end{remark}

The next result develops a specialized analysis of heavy ball for the case of
squared loss, which is not covered by the above theory. Its proof is quite
different from the restorativity-based approaches developed in
\citet{angelopoulos2025gradient} and used above. Instead, it relies on viewing
the evolution of heavy ball iterates for squared loss as a linear dynamical
system.

\begin{proposition}
\label{prop:quadratic-hb}
Consider heavy ball \eqref{eq:hb} with constant step sizes $\eta_t = \eta > 0$
for all $t \in \N$, $\beta \in [0,1)$, and arbitrary initialization $\theta_1
\in \R$. Let \smash{$\ell_t(\theta) = \frac12 (y_t-\theta)^2$} with $\abs{y_t}
\leq b$, for all $t \in \N$. Assume that \smash{$\eta < \frac{2(1+\beta)}
  {1-\beta}$}, and define
\[
A = \begin{bmatrix} \beta & 1-\beta \\ - \eta \beta & 1-\eta(1-\beta)
\end{bmatrix}.
\]
Write $\alpha = \norm{A}_2$ for the spectral norm of $A$, and $r = \rho(A)$ for
its spectral radius. Then, for all $t \in \N$,
\[
\abs{\theta_t} \leq \frac{\alpha+\beta}{1-r} \abs{\theta_1} +
b(1-\beta) \sqrt{1+\eta^2} \, \biggl(1 + \frac{\alpha+\beta}{(1-r)^2} \biggr).
\]
Therefore, for all $T \in \N$,
\[
\abs[\bigg]{\frac{1}{T} \sum_{t=1}^T g_t(\theta_t)}
  \leq \frac{\abs{\theta_1}}{\eta T} + \frac{1+\beta}{\eta(1-\beta) T} \biggl(
  \frac{\alpha+\beta}{1-r} \abs{\theta_1} + b(1-\beta) \sqrt{1+\eta^2} \,
  \Bigl(1 + \frac{\alpha+\beta}{(1-r)^2} \Bigr) \biggr).
\]
\end{proposition}

Similar to the discussion following \cref{cor:quantile-logistic}, a notable
application of \cref{prop:quadratic-hb} entails online adjustment of a black-box
prediction $f_t(x_t)$ of a response $z_t$, using heavy ball on the squared loss.
At each time $t$ we adjust the next prediction via
$f_{t+1}(x_{t+1})+\theta_{t+1}$, with $\theta_{t+1}$ learned by \eqref{eq:hb}.
In the language of the above proposition, each $y_t$ is replaced by the residual
$z_t-f_t(x_t)$, which we assume is bounded in magnitude by $b$. It is
straightforward to check that gradient equilibrium \eqref{eq:geq} in this
setting means debiasing the predictor:
\[
\frac{1}{T} \sum_{t=1}^T (f_t(x_t) + \theta_t) - \frac{1}{T} \sum_{t=1}^T z_t
\to 0,
\]
and \cref{sec:exp} presents an application of this type.

%% file: section/4-smoothness.tex
\section{Smoothness}
\label{sec:smoothness}

In this section, we examine the way that momentum methods alter the smoothness
properties of the corresponding output iterate sequence. As a smoothness penalty
(measure of nonsmoothness), we consider
\begin{equation}
\label{eq:smooth-penalty}
\frac{1}{T-2} \sum_{t=2}^{T-1} \norm{(\Delta^{\!2} \theta)_{t-1}}_2 = \frac{1}{T-2}
\sum_{t=2}^{T-1} \norm{\theta_{t+1} - 2\theta_t + \theta_{t-1}}_2.
\end{equation}
Here each $(\Delta^{\!2} \theta)_{t-1} = \theta_{t+1} - 2\theta_t +
\theta_{t-1}$ is the second difference of the sequence centered at $t$, which
may be seen as the discrete second derivative of the sequence at $t$. We note
that \eqref{eq:smooth-penalty} is a vector analog of discrete second-order total
variation, which underlies trend filtering methods in nonparametric regression
\citep{kim2009trend, tibshirani2014adaptive}. (\cref{app:path-norm} provides
further motivation for this choice of smoothness penalty.)

\subsection{Smoothness of heavy ball paths}

Below we give a sharp result for heavy ball. This is a consequence of a more
general result proved in the appendix, which also translates into a sharp result
for Nesterov momentum.

\begin{corollary}
\label{cor:hb-smooth}
Consider heavy ball \eqref{eq:hb} with positive step sizes $(\eta_t)_{t \in
\N}$, $\beta \in [0,1)$, and arbitrary initialization $\theta_1 \in \R^d$. Fix
any $T \geq 3$. If \smash{$\max_{t \leq T-1} \eta_t\norm{g_t(\theta_t)}_2 = 0$},
then the smoothness penalty is zero. Otherwise,
\begin{align*}
\frac{1}{T-2}\sum_{t=2}^{T-1} \norm{(\Delta^{\!2} \theta)_{t-1}}_2
&\leq \frac{2(1-\beta)}{1+\beta}
\max_{t \leq T-1} \eta_t\norm{g_t(\theta_t)}_2 +
\frac{\norm{\theta_2-\theta_1}_2^2-\norm{\theta_T-\theta_{T-1}}_2^2}
{2(T-2)\max_{t \leq T-1} \eta_t\norm{g_t(\theta_t)}_2} \\
&\leq \biggl(\frac{2(1-\beta)}{1+\beta} + \frac{(1-\beta)^2}{2(T-2)}\biggr)
\max_{t \leq T-1} \eta_t\norm{g_t(\theta_t)}_2.
\end{align*}
The first inequality is tight in the sense that equality is achieved by at least
one problem instance.
\end{corollary}

For heavy ball, the factor in parentheses in the upper bound of
\cref{cor:hb-smooth} decreases with $\beta$, suggesting that iterate paths are
more jagged for gradient descent ($\beta = 0$) and smoother as $\beta$
increases. Importantly, this is indeed what we observe in practice, as studied
in \cref{sec:exp}. (Note that the term \smash{$\max_{t \leq T-1}
\eta_t\norm{g_t(\theta_t)}_2$} may also implicitly depend on $\beta$ through the
iterates; however, the influence of $\beta$ on the gradient norms seems
negligible empirically.)

We emphasize again that the particular parametrizations we adopt for heavy ball
\eqref{eq:hb} and Nesterov \eqref{eq:nesterov} seem crucial to our smoothness
analyses; alternative versions of these methods may not lead to the same
conclusions. Furthermore, in \cref{app:post-hoc}, we give an equivalent
perspective on heavy ball as a special type of post-hoc smoothing. This helps
guide intuition about its smoothing capabilities (we also contrast it to other
post-hoc smoothing methods that do not preserve gradient equilibrium).

\subsection{GEQ--smoothness Pareto frontier}
\label{sec:pareto-hb}

\begin{figure}[t]
\centering
\includegraphics{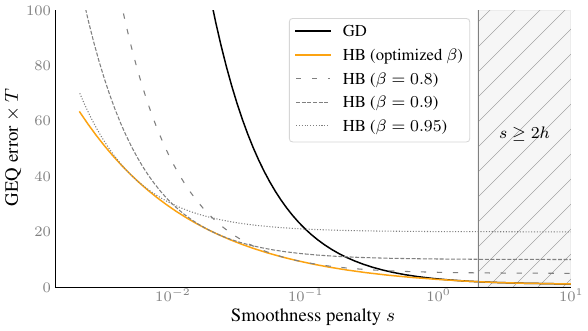}
\caption{Pareto frontiers for the gradient equilibrium and smoothness bounds,
  with $h = L = 1$. The GD frontier ($\beta = 0$, black) is $L (2h/s + 1)$,
  while the HB frontier (\textcolor{myorange}{orange}), optimized over $\beta$,
  is $2L \sqrt{2h/s}$ for $s \in (0,2h]$. The three gray curves show the bounds
  for HB with fixed $\beta$. The gray hatched region corresponds to $s \geq 2h$,
  where GD minimizes the bound and the two frontiers coincide.}
\label{fig:pareto-frontier}
\end{figure}

To conclude the theoretical part of the paper, we discuss the Pareto frontier
for the gradient equilibrium and smoothness bounds of gradient descent and heavy
ball. Consider the univariate case of \cref{cor:hb} under weak curvature (i), or
the multivariate case under positive curvature (iii) subject to its curvature
condition on $\eta$ and $\beta$, with initialization $\theta_1 = 0$, constant
step size $\eta_t = \eta > 0$, Lipschitz constant $L > 0$, and constant radius
$h > 0$. Since $1 + \beta \geq 1$, \cref{cor:hb-smooth} gives the smoothness
bound $2 \eta (1 - \beta) L + O(1/T)$. Setting $s > 0$ equal to the leading
term, we have $\eta = s/(2 (1 - \beta) L)$. Plugging this into the gradient
equilibrium bound in \cref{cor:hb} yields
$(L/T) (2h (1 - \beta)/s + 1/(1 - \beta))$ for all $T \geq 3$.

At each $s > 0$, minimizing this gradient equilibrium bound over $\beta$ gives
the Pareto frontier shown in \textcolor{myorange}{orange} in
\cref{fig:pareto-frontier}. The minimum is attained at
$\beta = 1 - \sqrt{s/(2h)} > 0$ for $0 < s < 2h$, and at $\beta = 0$ for
$s \geq 2h$. In the former case, the gradient equilibrium bound achieves a rate
of $2 \sqrt{2h/s}$, while for gradient descent ($\beta=0$) the rate is
$2h/s + 1$. These rates also apply, up to constant factors, to the squared
gradient equilibrium bound in case (ii).

Thus, optimizing momentum strictly improves the Pareto frontier over gradient
descent for $s \in (0,2h)$. Gradient descent minimizes this bound only for
$s \geq 2h$: at $\beta = 0$, this is equivalent to $\eta \geq h/L$, so one step
of gradient descent can suffice to switch between the minimum and maximum of
$(\norm{\theta_t}_2)_t$, potentially producing a highly jagged path. The regime
$s \geq 2h$ may therefore be undesirable. For $s < 2h$, heavy ball yields a
lower gradient equilibrium bound under the same smoothness guarantee. On a side
note, the dashed/dotted gray lines display how large $\beta$ can lead to high
gradient equilibrium error while still producing erratic, highly nonsmooth
trajectories (large smoothness penalty).

%% file: section/5-empirical-results.tex
\section{Experiments}
\label{sec:exp}

We expand on the experiments in the introduction. All figures and results are
reproducible from the code available at
\url{https://github.com/jv-rv/momentum-gradient-equilibrium/}. We focus on heavy
ball (HB) in \eqref{eq:hb}, which, recall, encompasses gradient descent (GD) as
a special case (when $\beta = 0$). We study two problem settings. Calibration
with quantile loss is described in \cref{sec:cal-exp}, with results in
\cref{fig:iterates-quantile-loss, fig:gradients-quantile-loss,
fig:frontier-quantile-loss, fig:smoothness-quantile-loss}. Debiasing with
squared loss is described in \cref{sec:deb-exp}, with results in
\cref{fig:iterates-squared-loss, fig:gradients-squared-loss,
fig:frontier-squared-loss, fig:smoothness-squared-loss}. Additional results for
multigroup debiasing appear in \cref{app:exp}.

\subsection{Experimental setup: calibration}
\label{sec:cal-exp}

For quantile loss, we use data on the volatility of BlackBerry stock, following
\citet{gibbs2021adaptive,gibbs2024conformal}. We use daily opening prices from
1999 through 2025, for a total sequence length of $T=\num{5517}$. To define the
response $z_t$ at time $t$, we first calculate simple daily returns $r_t = (p_t
- p_{t-1}) / p_{t-1}$, where $p_t$ is the price of BlackBerry at time $t$, and
then we set $z_t = r_t^2$. For the forecast \smash{$\hat\sigma_t^2$} at $t$, we
use a one-step-ahead volatility forecast from a GARCH(1,1) model fit to the most
recent \num{1250} observations \citep{bollerslev1986generalized}. Around each
forecast \smash{$\hat\sigma_t^2 \in (0,\infty)$}, we define a parametrized
prediction interval:
\[
C_t(\theta) = \biggl\{ z : \frac{\abs{z - \hat\sigma_t^2}}{\hat\sigma_t^2} \leq
\hat{Q}^t_{1-\theta} \biggr\} = [ (1 - \hat{Q}^t_{1-\theta}) \hat\sigma_t^2,
(1 + \hat{Q}^t_{1-\theta}) \hat\sigma_t^2 ],
\]
where \smash{$\hat{Q}^t_{1-\theta}$} is the empirical $(1-\theta)$-quantile of
the most recent relative errors prior to $t$ (up to \num{1250}), i.e.,
\smash{$\abs{z_s - \hat\sigma_s^2}/\hat\sigma_s^2$}. Following
\citet{gibbs2024conformal}, we set $\hat{Q}^t_q = \infty$ for $q \geq 1$ and
$\hat{Q}^t_q = -\infty$ for $q \leq 0$. To calibrate these prediction intervals
toward $1-\alpha$ sequential coverage, where $\alpha \in [0,1]$ is an arbitrary
user-chosen error level, we perform the following. At each time $t$, we set the
next prediction interval to be $C_{t+1}(\theta_{t+1})$, where $\theta_{t+1}$ is
given by a heavy ball update \eqref{eq:hb} based on the quantile loss of level
$\alpha$:
$\ell_t(\theta) = \alpha \abs{a_t-\theta} \, \mathbb{I}\{a_t\geq\theta\}
+ (1-\alpha) \abs{a_t-\theta} \, \mathbb{I}\{a_t<\theta\}$, with
$a_t = \sup\{a \in
[0,1] : z_t \in C_t(a)\}$ defined as the supremum of the error levels at which
the last prediction interval would have covered $z_t$. This approach is detailed
in \cref{alg:calibration}, which we run with $\alpha = 0.1$.

\begin{algorithm}
\caption{Calibrating black-box volatility forecasts}
\label{alg:calibration}
\KwIn{\justifying
  Responses $(z_t)_{t=1}^T$; predictions \smash{$(\hat\sigma_t^2)_{t=1}^{T+1}$};
  error level $\alpha \in [0,1]$; step size $\eta>0$; momentum $\beta\in[0,1)$}
Initialize $\theta_0=\theta_1=\alpha$\;
\For{$t = 1, \ldots, T$:}{
  Reveal last response: $z_t$\;
  Record gradient: $g_t = \mathbb{I}\{z_t \notin C_t(\theta_t)\} - \alpha$\;
  Perform update: $\theta_{t+1} = \theta_t + \beta (\theta_t-\theta_{t-1}) -
  \eta (1-\beta) g_t$\;
  Record next interval: $C_{t+1}(\theta_{t+1}) = [(1 -
  \hat{Q}^{t+1}_{1-\theta_{t+1}}) \hat\sigma_{t+1}^2, (1 +
  \hat{Q}^{t+1}_{1-\theta_{t+1}})\hat\sigma_{t+1}^2]$
}
\end{algorithm}

\subsection{Experimental setup: debiasing}
\label{sec:deb-exp}

For squared loss, we use hourly inflow observations at Shasta Reservoir for
February through June of each year from 2017 to 2025, from the California Data
Exchange Center, operated by the California Department of Water Resources
\citeyearpar{california2026cdec}. The total sequence length is $T=\num{32407}$
(we truncate our time-series plots to $400 < T \leq \num{15000}$ to avoid visual
crowding). Dams and their impounded reservoirs are key to water supply, power
generation, flood control, and environmental management. Operators mainly
control outflow (e.g., releases to cities and farms), while nature drives inflow
(e.g., precipitation, snowmelt, and streamflow). A key goal is to predict inflow
to inform outflow decisions. During severe droughts in 2014, 2015, and 2021,
forecast errors left Shasta Reservoir with insufficient water, causing
environmental harm \citep{null2022storing} and suggesting that forecasts were
biased upward. Had positive and negative forecast errors canceled out instead,
the damage could perhaps have been avoided.

For our experiments, we use hourly forecasts of full natural flow from NOAA's
California-Nevada River Forecast Center \citeyearpar{californianevada2026cnrfc}
to predict actual inflow, a goal similar to that of
\citet{georgakakos2014ensemble}. Each valid response (hourly inflow) is covered
by multiple forecasts made at different lead times. Among those with a recorded
modification time before the hour being forecast, we use only the most recent
one (the shortest lead time), where skill is highest on average. Even in this
favorable setting, a systematic bias relative to actual inflow persists,
motivating debiasing. This proceeds as follows. At each time $t$, we additively
adjust the next forecast $f_{t+1}$ to be $f_{t+1}+\theta_{t+1}$, where
$\theta_{t+1}$ is given by a heavy ball update \eqref{eq:hb} based on the
squared loss of the forecast residual $z_t-f_t$, i.e., \smash{$\ell_t(\theta) =
\frac12 (z_t-f_t-\theta)^2$}. This approach is detailed in \cref{alg:debiasing}.

\begin{algorithm}
\caption{Debiasing black-box predictions}
\label{alg:debiasing}
\KwIn{Responses $(z_t)_{t=1}^T$; predictions $(f_t)_{t=1}^{T+1}$; step size
  $\eta>0$; momentum $\beta\in[0,1)$}
Initialize $\theta_0=\theta_1=0$\;
\For{$t = 1, \ldots, T$:}{
  Reveal last response: $z_t$\;
  Record gradient: $g_t = f_t + \theta_t - z_t$\;
  Perform update: $\theta_{t+1} = \theta_t + \beta (\theta_t-\theta_{t-1}) -
  \eta (1-\beta) g_t$\;
  Record next forecast: $f_{t+1}+\theta_{t+1}$
}
\end{algorithm}

\begin{figure}[t]
\centering
\includegraphics{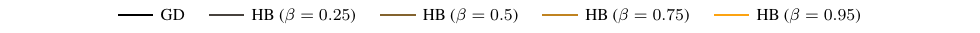}

\begin{subfigure}[b]{0.5\textwidth}
\centering
\includegraphics{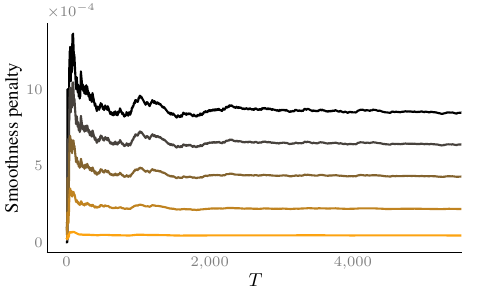}
\caption{Quantile loss: smoothness penalty.}
\label{fig:smoothness-quantile-loss}
\end{subfigure}%
\begin{subfigure}[b]{0.5\textwidth}
\centering
\includegraphics{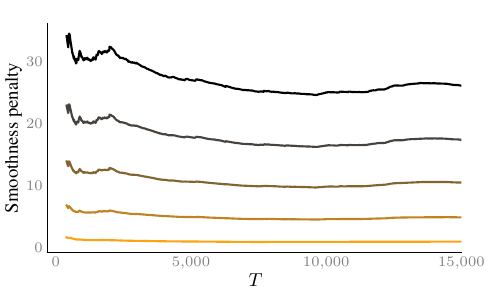}
\caption{Squared loss: smoothness penalty.}
\label{fig:smoothness-squared-loss}
\end{subfigure}

\medskip
\begin{subfigure}[b]{0.5\textwidth}
\centering
\includegraphics{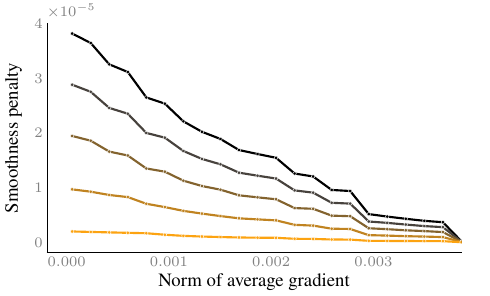}
\caption{Quantile loss: smoothness versus GEQ error.}
\label{fig:frontier-quantile-loss}
\end{subfigure}%
\begin{subfigure}[b]{0.5\textwidth}
\centering
\includegraphics{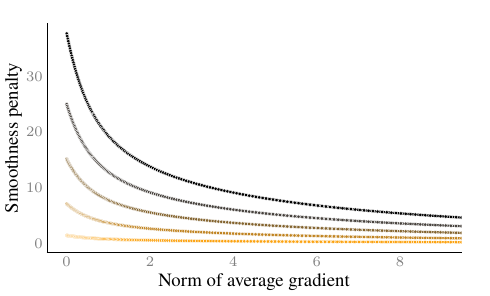}
\caption{Squared loss: smoothness versus GEQ error.}
\label{fig:frontier-squared-loss}
\end{subfigure}
\caption{Smoothness penalty over time for $\eta = 0.005$ (top), and smoothness
  penalty versus gradient equilibrium error at the final iteration $T$ as $\eta$
  varies (bottom), for quantile loss (left) and squared loss (right). Increasing
  the momentum parameter $\beta$ leads to smoother paths, and for larger
  $\beta$, HB achieves a much lower smoothness penalty at comparable gradient
  equilibrium error.}
\label{fig:experiments}
\end{figure}

\subsection{Results}

\cref{fig:smoothness-quantile-loss, fig:smoothness-squared-loss} track the
smoothness penalty \eqref{eq:smooth-penalty} over time $T$, with $\eta = 0.005$
and $\beta$ varying. These plots isolate the influence of the momentum term
$m_t$: each curve represents a different weighting of the newest gradient
$g_t(\theta_t)$ and the previous momentum term $m_{t-1}$, which are the two
ingredients used to define $m_t$. They also complement \cref{fig:introduction}
in the introduction, which displayed the iterates and norm of the average
gradient for the same experiments. We can clearly see from these plots that HB
obtains a lower smoothness penalty for larger $\beta$, consistent with the bound
in \cref{cor:hb-smooth}.

\cref{fig:frontier-quantile-loss, fig:frontier-squared-loss} display the
smoothness penalty versus the norm of the average gradient as we vary both
$\eta$ and $\beta$, and compute metrics at the final iteration $T$ for each
dataset. Each color corresponds to a value of $\beta$, and each dot a value of
$\eta$. The results show that GD ($\beta=0$) cannot be effectively smoothed by
simply decreasing $\eta$, because this inflates its gradient equilibrium error.
Meanwhile, HB ($\beta>0$) unlocks previously inaccessible dynamics, seeing as it
achieves much smoother paths at comparable gradient equilibrium error to GD.
Transparently, HB dominates GD with respect to the quantitative tradeoff traced
out between smoothness and gradient equilibrium error.

%% file: section/6-discussion.tex
\section{Discussion}

In this paper, we showed that momentum methods such as heavy ball and Nesterov
momentum effectively navigate the tradeoff between gradient equilibrium error
and iterate smoothness. These methods balance the two criteria in a manner that
cannot be recreated using ordinary gradient descent: when the step sizes are
tuned to produce matching gradient equilibrium errors, gradient descent
generally produces more jagged iterate paths than heavy ball or Nesterov, both
in theory and in practice.

An important related question is how smoothness affects downstream decisions. As
noted in the introduction, an erratic sequence of predicted returns could induce
frequent changes in portfolio holdings, potentially incurring large transaction
costs. Similarly, in the setting of our debiasing experiment, opening and
closing dam gates can incur operating costs. An erratic sequence of inflow
predictions could induce frequent changes in outflow decisions and increase
costs. These examples provide conceptual motivation for studying smoothness as a
desirable property. As in many methodological developments, we began by
formulating abstractions, and our theoretical and empirical results should lay
the groundwork for studying more precisely how smoothness relates to specific
downstream decisions.

We close by touching on some other directions that may be of interest for future
work. One is to extend the theory of gradient equilibrium for generalized
momentum methods to more broadly accommodate non-Lipschitz losses; this paper
only handled the special combination of squared loss and the heavy ball method.
Another would be to study the application of momentum in lazy gradient updates
(also called follow the regularized leader), which turn out to be critical in
multilevel calibration problems \citep{ding2025calibrated}. It would also be
interesting to determine whether heavy ball is optimal for the tradeoff
investigated in this paper; heavy ball appears to outperform Nesterov momentum
in terms of sharp upper bounds on the second-order smoothness penalty we defined
(compare \cref{cor:hb-smooth, cor:nesterov-smooth}); however, another method
might be better still. Finally, we view this work as a first step toward
analyzing gradient equilibrium for online methods which combine momentum and
adaptive preconditioning, such as Adam \citep{kingma2015adam}, which has become
a cornerstone of deep learning optimization toolkits.

%% file: section/7-appendix.tex
The appendices are organized as follows. \cref{app:geq} provides proofs of
results in \cref{sec:geq}, along with additional results for Nesterov momentum.
\cref{app:smoothness} provides proofs of results in \cref{sec:smoothness}, and
again additional results for Nesterov momentum. \cref{app:exp} contains
additional empirical analysis.

\section[Proofs for gradient equilibrium and related results]
{Proofs for \cref{sec:geq} and related results}
\label{app:geq}

\subsection{Abel transformation}

\begin{lemma}
\label{lem:recast}
Let $a_t \in \R^d$ for $t=1,\ldots,T+1$ and $\lambda_t \in \R$ for
$t=1,\ldots,T$. Then,
\[
\sum_{t=1}^T \lambda_t (a_t - a_{t+1}) = \lambda_1 (a_1 - a_{T+1}) +
\sum_{t=2}^T (\lambda_t - \lambda_{t-1}) (a_t - a_{T+1}).
\]
\end{lemma}

\begin{proof}
\begin{align*}
\sum_{t=1}^T \lambda_t (a_t - a_{t+1})
&= \sum_{t=1}^T \lambda_t a_t - \sum_{t=2}^{T+1} \lambda_{t-1} a_t \\
&= \lambda_1 a_1 + \sum_{t=2}^T \lambda_t a_t - \sum_{t=2}^T
\lambda_{t-1} a_t  - \lambda_T  a_{T+1} \\
&= \lambda_1 a_1 - \lambda_T a_{T+1} + \sum_{t=2}^T
(\lambda_t - \lambda_{t-1}) a_t \\
&= \lambda_1 (a_1 - a_{T+1}) + (\lambda_1 - \lambda_T) a_{T+1} +
\sum_{t=2}^T (\lambda_t - \lambda_{t-1}) a_t \\
&= \lambda_1 (a_1 - a_{T+1}) + \sum_{t=2}^T (\lambda_{t-1} - \lambda_t)
a_{T+1} + \sum_{t=2}^T (\lambda_t - \lambda_{t-1}) a_t \\
&= \lambda_1 (a_1 - a_{T+1}) + \sum_{t=2}^T (\lambda_t - \lambda_{t-1})
(a_t - a_{T+1}).\qedhere
\end{align*}
\end{proof}

\subsection{Bound between auxiliary and original iterates}

\begin{lemma}
\label{lem:aux}
Under \cref{ass:bound-stepsize-scheme}, the original and auxiliary iterates
satisfy, for any $t \in \N$,
\[
\norm{\theta_t - \aux_t}_2 \leq \delta \max_{s \leq t-1} \norm{g_s(\theta_s)}_2.
\]
\end{lemma}

\begin{proof}
For $t = 1$, the result follows from $\aux_1 = \theta_1$ and the convention that
the maximum over the empty set is zero. We may therefore assume that $t \geq 2$.
Using the unrolled expressions of $\theta_t$ and $\aux_t$, we have
\begin{align*}
\theta_t - \aux_t
&= \sum_{k=1}^{t-1} \rho_k g_k(\theta_k) - \sum_{k=1}^{t-1} \sum_{u=k}^{t-1}
  \gamma_{uk} g_k(\theta_k) \\
&= \sum_{k=1}^{t-1} \biggl(\rho_k - \sum_{u=k}^{t-1} \gamma_{uk} \biggr)
g_k(\theta_k) \\
&= \sum_{k=1}^{t-1} \sum_{u=t}^{\infty} \gamma_{uk} g_k(\theta_k),
\end{align*}
where the last line follows from the definition of $\rho_k$. Therefore,
\begin{align*}
\norm{\theta_t - \aux_t}_2
&\leq \sum_{k=1}^{t-1} \sum_{u=t}^{\infty} \gamma_{uk}
\norm{g_k(\theta_k)}_2 \\
&\leq \max_{s\leq t-1} \norm{g_s(\theta_s)}_2 \sum_{k=1}^{t-1}
\sum_{u=t}^{\infty} \gamma_{uk} \\
&\leq \delta \max_{s\leq t-1} \norm{g_s(\theta_s)}_2.\qedhere
\end{align*}
\end{proof}

\subsection[Proof of the lemma on the average gradient]
{Proof of \cref{lem:general-redux}}

Set $a_0 = 0$ and $a_t = 1/\rho_t$ for all $t \in \N$. By the definition of the
auxiliary sequence,
\[
g_t(\theta_t) = a_t (\aux_t - \aux_{t+1}).
\]
Summing and applying \cref{lem:recast} yields
\[
\frac{1}{T} \sum_{t=1}^T g_t(\theta_t)
  = \frac{1}{T} \sum_{t=1}^T (a_t - a_{t-1}) (\aux_t - \aux_{T+1}).
\]

Since $\aux_1 = \theta_1$, $a_1 - a_0 = \rho_1^{-1}$, and $a_t - a_{t-1} =
\rho_t^{-1} - \rho_{t-1}^{-1}$ for $t \geq 2$, the triangle inequality gives
\[
\norm[\bigg]{\frac{1}{T} \sum_{t=1}^T g_t(\theta_t)}_2
  \leq \frac{\norm{\theta_1 - \aux_{T+1}}_2}{\rho_1 T} + \frac{1}{T}
  \sum_{t=2}^T \abs{\rho_t^{-1} - \rho_{t-1}^{-1}} \, \norm{\aux_t -
    \aux_{T+1}}_2.
\]
For all $t \leq T$,
\[
\norm{\aux_t - \aux_{T+1}}_2
  \leq 2 \max_{t \leq T+1} \norm{\aux_t}_2.
\]
Combining the last two displays proves the first inequality. Moreover,
$\norm{\theta_1 - \aux_{T+1}}_2 \leq 2 \max_{t \leq T+1} \norm{\aux_t}_2$, so
\[
\norm[\bigg]{\frac{1}{T} \sum_{t=1}^T g_t(\theta_t)}_2
  \leq \frac{2 \max_{t \leq T+1} \norm{\aux_t}_2}{T} \biggl(\rho_1^{-1} +
  \sum_{t=2}^T \abs{\rho_t^{-1} - \rho_{t-1}^{-1}} \biggr),
\]
which proves the second inequality.

\subsection{Bound between momentum and gradient descent paths}

One can also use the auxiliary sequence to show that the path traced by the
generalized momentum method is close to a non-momentum counterpart which makes
pure gradient updates. This is not used in the main paper, but is included below
because it may be of standalone interest.

\begin{lemma}
\label{lem:close-paths}
Consider \eqref{eq:gmm} with arbitrary initialization $\theta_1 \in \R^d$.
Define \smash{$\gd_1 = \theta_1$}, and consider gradient descent updates
\smash{$\gd_{t+1} = \gd_t - \rho_t g_t(\gd_t)$}. Under
\cref{ass:bound-stepsize-scheme}, and assuming each $\ell_t$ is $L$-Lipschitz,
it holds for any $t \in \N$ that
\[
\norm{\theta_t - \gd_t}_2 \leq \delta L + 2L \sum_{k=1}^{t-1} \abs{\rho_k}.
\]
\end{lemma}

\begin{proof}
Unraveling the auxiliary update rule, we have \smash{$\aux_t = \theta_1 -
\sum_{k=1}^{t-1} \rho_k g_k(\theta_k)$}. Similarly, unraveling the gradient
descent rule, we have \smash{$\gd_t = \theta_1 - \sum_{k=1}^{t-1} \rho_k
  g_k(\gd_k)$}, which yields
\[
\aux_t - \gd_t = \sum_{k=1}^{t-1} \rho_k (g_k(\gd_k) - g_k(\theta_k)).
\]
Therefore, by the triangle inequality and $L$-Lipschitzness,
\[
\norm{\aux_t - \gd_t}_2
\leq \sum_{k=1}^{t-1} \abs{\rho_k} \, \norm{g_k(\gd_k) - g_k(\theta_k)}_2
\leq \sum_{k=1}^{t-1} \abs{\rho_k} (\norm{g_k(\gd_k)}_2 + \norm{g_k(\theta_k)}_2)
\leq 2L \sum_{k=1}^{t-1} \abs{\rho_k}.
\]
By \cref{lem:aux} and $L$-Lipschitzness, $\norm{\theta_t - \aux_t}_2 \leq \delta
L$, and the desired result follows from the triangle inequality.
\end{proof}

\subsection[Proof of the proposition in one dimension]
{Proof of \cref{prop:weak-curvature-1d}}

Fix $T \in \N$, and define $B_T = \max\{\abs{\theta_1}, h_T + \delta L\} + L
\max_{t \leq T} \rho_t$. We prove by induction that $\abs{\aux_t} \leq B_T$ for
all $t \leq T+1$. The base case follows from $\abs{\aux_1} = \abs{\theta_1}
\leq B_T$. For the inductive step, fix $t \leq T$, assume that $\abs{\aux_t}
\leq B_T$, and consider two cases. If $\abs{\aux_t} \leq h_t + \delta L$, then
\[
\abs{\aux_{t+1}}
  \leq h_t + \delta L + \rho_t L
  \leq h_T + \delta L + L \max_{s \leq T} \rho_s
  \leq B_T.
\]
Otherwise, $\abs{\aux_t} > h_t + \delta L$. As $\abs{\aux_t - \theta_t} \leq
\delta L$, we have $\abs{\theta_t} \geq \abs{\aux_t} - \delta L > h_t$. By
squaring $\abs{\aux_t - \theta_t} < \abs{\aux_t}$, we see that $\aux_t^2 +
\theta_t^2 - 2\aux_t \theta_t < \aux_t^2$. Rearranging, we see that $\aux_t$ and
$\theta_t$ have the same sign: $\aux_t \theta_t > \theta_t^2 / 2 \geq
0$. Together with the restorativity condition, this implies $\aux_t
g_t(\theta_t) \geq 0$. Therefore,
\[
\abs{\aux_{t+1}}
  = \abs[\big]{\abs{\aux_t} - \rho_t \abs{g_t(\theta_t)}}
  \leq \max\{\abs{\aux_t}, \rho_t L\}
  \leq B_T.
\]
This completes the inductive proof, and establishes the bound on auxiliary
iterates.

\subsection[Proof of the corollary for quantile and logistic losses]
{Proof of \cref{cor:quantile-logistic}}

Quantile loss is $1$-Lipschitz, and generalized logistic loss is
$(b-a)$-Lipschitz. Corollaries 10 and 12 in \citet{angelopoulos2025gradient}
show that both losses are restorative. If $(\rho_t)_{t \in \N}$ is
nonincreasing, then the sum involving $\abs{\rho_t^{-1} - \rho_{t-1}^{-1}}$ in
\cref{prop:weak-curvature-1d} telescopes to $\rho_T^{-1} - \rho_1^{-1}$, and
$\max_{t \leq T} \rho_t = \rho_1$. \cref{prop:weak-curvature-1d} gives the
$O((\rho_T T)^{-1})$ bound and, since $\gamma_{TT} \leq \rho_T$, also the
$O((\gamma_{TT} T)^{-1})$ bound whenever $\gamma_{TT} > 0$. For constant step
sizes $\gamma_{tt} = \eta > 0$, this bound is $O(1/T)$.

\subsection[Proof of the proposition with weak curvature]
{Proof of \cref{prop:weak-curvature}}

Observe that
\begin{align*}
\norm{\aux_{T+1}}_2^2
&= \norm{\aux_T - \rho_T g_T(\theta_T)}_2^2 \\
&= \norm{\aux_T}_2^2 - 2 \rho_T g_T(\theta_T)^\T \aux_T + \rho_T^2
  \norm{g_T(\theta_T)}_2^2 \\
&\leq \norm{\aux_T}_2^2 - 2\rho_T g_T(\theta_T)^\T (\aux_T - \theta_T) -
2\rho_T g_T(\theta_T)^\T \theta_T + \rho_T^2 L^2 \\
&\leq \norm{\aux_T}_2^2 + 2\rho_T L \norm{\aux_T - \theta_T}_2 -
2\rho_T g_T(\theta_T)^\T \theta_T + \rho_T^2 L^2 \\
&\leq \norm{\aux_T}_2^2 + 2\rho_T \delta L^2 - 2\rho_T g_T(\theta_T)^\T \theta_T
+ \rho_T^2 L^2 \\
&\leq \norm{\aux_T}_2^2 + 2\rho_T (\delta L^2 + Lh_T) + \rho_T^2 L^2,
\end{align*}
where the second-to-last inequality follows from \cref{lem:aux} and the last
inequality follows by $-g_T(\theta_T)^\T \theta_T \leq h_T L$, which can be
argued in cases: if $\norm{\theta_T}_2 \leq h_T$, this follows from
Cauchy--Schwarz, whereas if $\norm{\theta_T}_2 > h_T$, we have
$-g_T(\theta_T)^\T \theta_T \leq 0$ by restorativity. Unrolling, and recalling
$\aux_1 = \theta_1$,
\[
\norm{\aux_{T+1}}_2^2 \leq \norm{\theta_1}_2^2 + L^2 \sum_{t=1}^T \rho_t^2 + 2L
\sum_{t=1}^T (h_t + \delta L) \rho_t.
\]
Since the summands are nonnegative, the same argument bounds $\norm{\aux_s}_2^2$
by the right-hand side above for every $s \leq T+1$, which establishes the bound
on auxiliary iterates.

\subsection[Proof of the proposition with positive curvature]
{Proof of \cref{prop:positive-curvature}}

Fix $T \in \N$, and define $B_T = \max\{\norm{\theta_1}_2, h_T + \delta L\} + L
\max_{t \leq T} \rho_t$. As in the proof of \cref{prop:weak-curvature-1d}, we
prove by induction that $\norm{\aux_t}_2 \leq B_T$ for all $t \leq T+1$. The
base case follows from $\norm{\aux_1}_2 = \norm{\theta_1}_2 \leq B_T$. For the
inductive step, fix $t \leq T$, assume that $\norm{\aux_t}_2 \leq B_T$, and
consider two cases. If $\norm{\aux_t}_2 \leq h_t + \delta L$, then
\[
\norm{\aux_{t+1}}_2
  \leq h_t + \delta L + \rho_t L
  \leq h_T + \delta L + L \max_{s \leq T} \rho_s
  \leq B_T.
\]
Otherwise, $\norm{\aux_t}_2 > h_t + \delta L$. As $\norm{\aux_t - \theta_t}_2
\leq \delta L$, we have $\norm{\theta_t}_2 \geq \norm{\aux_t}_2 - \delta L >
h_t$. Expanding and bounding terms as in the proof of
\cref{prop:weak-curvature}, $\norm{\aux_{t+1}}_2^2 \leq \norm{\aux_t}_2^2 +
2\rho_t \delta L^2 - 2\rho_t g_t(\theta_t)^\T \theta_t + \rho_t^2 L^2$. Write
the last three terms as
\[
2\rho_t \delta L^2 - 2\rho_t g_t(\theta_t)^\T \theta_t + \rho_t^2 L^2
  = 2\rho_t((\delta + \rho_t/2) L^2 - g_t(\theta_t)^\T \theta_t).
\]
By the restorativity condition, this expression is nonpositive. Therefore,
$\norm{\aux_{t+1}}_2 \leq \norm{\aux_t}_2 \leq B_T$. This completes the
inductive proof, and establishes the bound on auxiliary iterates.

\subsection[Proof of the lemma for heavy ball]
{Proof of \cref{lem:hb}}

Since $\eta_k > 0$ and $\beta \in [0,1)$, the coefficients $\gamma_{tk}$ are
nonnegative. To verify the remaining condition in
\cref{ass:bound-stepsize-scheme}, note that
\begin{align*}
\sum_{k=1}^{t-1} \sum_{u=t}^\infty \gamma_{uk}
  &= \sum_{k=1}^{t-1} \eta_k (1-\beta) \sum_{u=t}^\infty \beta^{u-k} \\
  &= \sum_{k=1}^{t-1} \eta_k \beta^{t-k} \\
  &\leq \eta_1 \sum_{j=1}^{t-1} \beta^j \\
  &\leq \eta_1 \sum_{j=1}^\infty \beta^j \\
  &= \frac{\eta_1\beta}{1-\beta},
\end{align*}
i.e., we may take $\delta = \eta_1\beta/(1-\beta)$. Moreover,
\[
\rho_k = \sum_{u=k}^\infty \eta_k (1-\beta) \beta^{u-k} = \eta_k.
\]

\subsection[Proof of the corollary for heavy ball]
{Proof of \cref{cor:hb}}

Since $\rho_t = \eta_t$ is nonincreasing,
\[
\eta_1^{-1} + \sum_{t=2}^T \abs{\eta_t^{-1} - \eta_{t-1}^{-1}}
  = \eta_T^{-1}.
\]
Combining this identity and \cref{lem:general-redux, lem:hb} with
\cref{prop:weak-curvature-1d, prop:weak-curvature, prop:positive-curvature}
proves the result.

\subsection[Proof of the proposition for squared loss]
{Proof of \cref{prop:quadratic-hb}}

As noted below the definition of \eqref{eq:hb}, in the case of constant step
sizes, the heavy ball updates may be written as
\begin{align*}
m_t &= \beta m_{t-1} + (1-\beta) g_t(\theta_t), \\
\theta_{t+1} &= \theta_t - \eta m_t.
\end{align*}
In the current case, these recurrences become
\begin{align*}
m_t &= \beta m_{t-1} + (1-\beta) (\theta_t - y_t), \\
\theta_{t+1} &= \theta_t - \eta m_t.
\end{align*}
Define $x_{t+1} = (m_t, \theta_{t+1})^\T \in \R^2$. Then, we may rewrite the
above as
\begin{align*}
x_{t+1,1} &= \beta x_{t1} + (1-\beta) x_{t2} - (1-\beta) y_t, \\
x_{t+1,2} &= - \eta \beta x_{t1} + (1 - \eta (1-\beta)) x_{t2} + \eta (1-\beta)
  y_t.
\end{align*}

Defining $A$ as in the proposition statement and $c = (\beta-1, \eta
(1-\beta))^\T \in \R^2$, we can represent this more compactly as
\[
x_{t+1} = A x_t + c y_t.
\]
Unrolling,
\[
x_{t+1} = A^t x_1 + \sum_{k=0}^{t-1} A^k c y_{t-k}.
\]
Since $\norm{x_1}_2 = \abs{\theta_1}$, we can bound $\abs{\theta_{t+1}} \leq
\norm{x_{t+1}}_2$, and the latter by
\[
\norm{x_{t+1}}_2 \leq \norm{A^t}_2 \abs{\theta_1} + b \norm{c}_2 \sum_{k=0}^{t-1} \norm{A^k}_2,
\]
where we have used the triangle inequality and $\abs{y_t} \leq b$ for all $t \in
\N$. We further upper bound the right-hand side by
\[
\norm{x_{t+1}}_2 \leq \abs{\theta_1} \sup_{k \in \N} \norm{A^k}_2 + b
\norm{c}_2 \sum_{k=0}^\infty \norm{A^k}_2.
\]
By \cref{lem:spectral-radius}, we know that $r < 1$ provided \smash{$\eta <
\frac{2(1+\beta)}{1-\beta}$}. In this case, by \cref{lem:spectral-sup}, we have
\[
\sup_{k \in \N} \norm{A^k}_2 \leq \frac{\alpha+\beta}{1-r}.
\]
Moreover, \cref{lem:spectral-sum} gives
\[
\sum_{k=0}^\infty \norm{A^k}_2 \leq 1+\frac{\alpha+\beta}{(1-r)^2}.
\]
Putting these together, along with the simple calculation \smash{$\norm{c}_2 =
(1-\beta)\sqrt{1+\eta^2}$}, proves the iterate bound in the proposition. The
case $t = 1$ follows from $\alpha \geq 1-\beta$, which implies
$(\alpha+\beta)/(1-r) \geq 1$.

Finally, summing \eqref{eq:hb} gives
\[
\frac{1}{T} \sum_{t=1}^T g_t(\theta_t)
  = \frac{(1-\beta)\theta_1 + \beta \theta_T - \theta_{T+1}}{\eta (1-\beta) T}.
\]
The bound on the average gradient then follows from the triangle inequality and
the iterate bound.

\subsection{Auxiliary heavy ball lemmas}

\begin{lemma}
\label{lem:spectral-radius}
For $A$ as defined above, with $\eta>0$ and $\beta \in [0,1)$, we have
$\rho(A) < 1$ if and only if \smash{$\eta < \frac{2(1+\beta)}{1-\beta}$}.
\end{lemma}

\begin{proof}
Let $\tau = \tr(A) = \beta + 1-\eta(1-\beta)$. Note
\[
\det(A) = \beta (1-\eta(1-\beta)) + (1-\beta) \eta\beta = \beta.
\]
Thus the characteristic polynomial of $A$ is
\[
p(\lambda) = \lambda^2- \tau\lambda +\beta.
\]
Let $\lambda_1,\lambda_2$ be the roots of $p$, i.e., eigenvalues of $A$. These
roots lie strictly inside the unit disk if and only if $\abs{\tau} < 1+\beta$,
i.e., $-1-\beta < \tau < 1+\beta$. Substituting $\tau = 1+\beta -
\eta(1-\beta)$, note that the upper inequality becomes $-\eta(1-\beta) < 0$,
which always holds. The lower inequality becomes $\eta(1-\beta) < 2(1+\beta)$,
which after rearrangement gives the stated restriction on $\eta$.
\end{proof}

\begin{lemma}
\label{lem:spectral-sup}
For $A$ as defined above, with \smash{$\eta \in (0,
  \frac{2(1+\beta)}{1-\beta})$} and $\beta \in [0,1)$, it holds for all $k \in
\N$ that
\[
\norm{A^k}_2 \leq \frac{\alpha+\beta}{1-r},
\]
where, recall, $\alpha = \norm{A}_2$ and $r = \rho(A)$.
\end{lemma}

\begin{proof}
By the Cayley--Hamilton theorem, $A^2-\tau A+\beta I=0$, with $\tau = \tr(A)$
and $\beta = \det(A)$ as calculated in the proof of \cref{lem:spectral-radius}.
Rearranged, this says that
\[
A^2=\tau A-\beta I.
\]
Now define $u_0 = 0$, $u_1 = 1$, and for $k \geq 1$,
\[
u_{k+1} = \tau u_k - \beta u_{k-1}.
\]
It is straightforward to verify by induction that for $k \geq 1$,
\[
A^k = u_kA-\beta u_{k-1} I.
\]
The base case $k = 1$ holds by the definitions of $u_0$ and $u_1$, and for the
inductive step, we can compute
\[
A^{k+1} = AA^k = A(u_kA-\beta u_{k-1}I),
\]
using the inductive hypothesis. Expanding the right-hand side and using the
identities for $A^2$ and $u_{k+1}$, we obtain
\[
A^{k+1} = u_kA^2-\beta u_{k-1}A = u_k(\tau A-\beta I) -\beta u_{k-1} A =
u_{k+1}A - \beta u_k I,
\]
which completes the induction.

Now with $\lambda_1,\lambda_2$ denoting the eigenvalues of $A$, the standard
solution to the second-order recurrence $u_{k+1} - \tau u_k + \beta u_{k-1} = 0$
is of the form
\[
u_k = a \lambda_1^k + b \lambda_2^k.
\]
Using the initial conditions, we learn from $u_0 = 0$ that $a+b=0$, and from
$u_1 = 1$ that $a (\lambda_1-\lambda_2) = 1$. So in the case of nonrepeated
roots $\lambda_1\neq\lambda_2$, we have $a = 1/(\lambda_1-\lambda_2)$ and $b =
-1/(\lambda_1-\lambda_2)$. Thus
\[
u_k = \frac{\lambda_1^k - \lambda_2^k}{\lambda_1 - \lambda_2}
= \sum_{j=0}^{k-1} \lambda_1^{k-1-j} \lambda_2^j.
\]
In the case of repeated roots, $\lambda_1=\lambda_2=\lambda$, one can check that
$u_k = k\lambda^{k-1}$ and the above representation is still valid. In either
case,
\[
\abs{u_k} \leq \sum_{j=0}^{k-1} \abs{\lambda_1}^{k-1-j} \abs{\lambda_2}^j \leq k
r^{k-1}.
\]
For $k = 1$, the desired result follows from $\alpha \leq (\alpha+\beta)/(1-r)$.
For all $k \in \N$ with $k \geq 2$, the representation for $A^k$ gives
\[
\norm{A^k}_2 \leq \abs{u_k} \, \norm{A}_2 + \beta \abs{u_{k-1}}
\leq k r^{k-1 }\alpha + (k-1) r^{k-2} \beta.
\]
Plugging in the simple bounds $k r^{k-1} \leq 1/(1-r)$ and $(k-1)r^{k-2} \leq
1/(1-r)$ proves the result.
\end{proof}

\begin{lemma}
\label{lem:spectral-sum}
For $A$ as defined above, with \smash{$\eta \in (0,
\frac{2(1+\beta)}{1-\beta})$} and $\beta \in [0,1)$, it holds that
\[
\sum_{k=0}^\infty \norm{A^k}_2 \leq 1+\frac{\alpha+\beta}{(1-r)^2},
\]
where, recall, $\alpha = \norm{A}_2$ and $r = \rho(A)$.
\end{lemma}

\begin{proof}
Using $A^0 = I$, $A^k = u_k A - \beta u_{k-1} I$ for all $k \in \N$, and the
estimates in the proof of \cref{lem:spectral-sup},
\begin{align*}
\sum_{k=0}^\infty \norm{A^k}_2
  &\leq 1 + \alpha \sum_{k=1}^\infty \abs{u_k} + \beta \sum_{k=1}^\infty \abs{u_{k-1}} \\
  &\leq 1 + \alpha \sum_{k=1}^\infty k r^{k-1} + \beta \sum_{k=2}^\infty (k-1)r^{k-2} \\
  &= 1 + \frac{\alpha+\beta}{(1-r)^2}.\qedhere
\end{align*}
\end{proof}

\subsection{Heavy ball for locally Lipschitz losses}
\label{app:swing}

\begin{figure}[t]
\centering
\begin{subfigure}[b]{0.5\textwidth}
\centering
\includegraphics{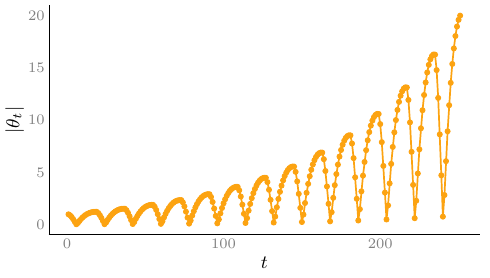}
\caption{$\beta = 0.9$.}
\label{fig:swing-heavy-ball}
\end{subfigure}%
\begin{subfigure}[b]{0.5\textwidth}
\centering
\includegraphics{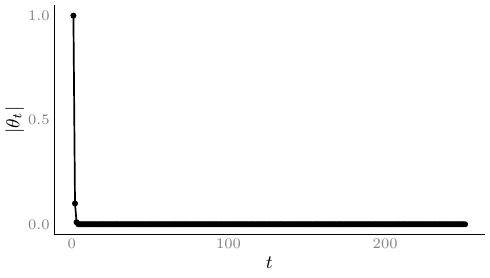}
\caption{$\beta = 0$.}
\label{fig:swing-gradient-descent}
\end{subfigure}
\caption{Heavy ball iterates for asymmetric squared losses.}
\label{fig:swing}
\end{figure}

We investigate heavy ball with $\beta=0.9$ and $\beta=0$ (i.e., gradient
descent). In each case we set $\eta = 0.9$, and we consider the following
asymmetric squared losses:
\begin{align*}
\ell_L(\theta) &= \frac{M}{2}\theta^2\mathbb{I}\{\theta \leq 0\} +
\frac{m}{2}\theta^2\mathbb{I}\{\theta > 0\}, \\
\ell_R(\theta) &= \frac{m}{2}\theta^2\mathbb{I}\{\theta \leq 0\} +
\frac{M}{2}\theta^2\mathbb{I}\{\theta > 0\},
\end{align*}
with $M = 1$ and $m = 0.1$. It is straightforward to check that each loss is
Lipschitz smooth with parameter $L=M$ and strongly convex with parameter
$\mu=m$. Furthermore, each is restorative with quadratic curvature, which means
\smash{$\ell'(\theta) \theta \geq \frac{\eta}{2} (\ell'(\theta))^2$} for large
enough $\abs{\theta}$. To see this, for $\ell_L$ and $\theta > 0$, quadratic
curvature requires
\[
m \theta^2 \geq \frac{\eta}{2} m^2 \theta^2,
\]
which holds as long as $\eta \leq 2/m$. For $\theta < 0$, it requires
\[
M \theta^2 \geq \frac{\eta}{2} M^2 \theta^2,
\]
which holds as long as $\eta \leq 2/M$. Together, these restrictions give $\eta
\leq \min\{2/m, 2/M\} = 2/M$, which clearly holds for $\eta = 0.9$ and $M = 1$.
The argument for $\ell_R$ is identical.

We initialize at $\theta_1 = 1$, and for each $t \in \N$, we set $\ell_t =
\ell_R$ if $m_{t-1} \geq 0$ and $\ell_t = \ell_L$ otherwise. In other words, the
larger quadratic coefficient is on the right when the previous momentum is
nonnegative and on the left otherwise. This induces a type of oscillatory
behavior in the iterates whenever $\beta \neq 0$ (nontrivial accumulation of
momentum). \cref{fig:swing} depicts the evolution of the iterates for the first
250 iterations; note the exponential growth of iterates for $\beta=0.9$ and the
exponential contraction for $\beta=0$.

\subsection{Nesterov momentum}
\label{app:nesterov}

We specialize the general theory to Nesterov momentum, which refers to the
auxiliary sequence in \eqref{eq:nesterov}. The auxiliary sequence itself is an
instance of \eqref{eq:gmm}; unrolling \eqref{eq:nesterov} gives
\[
(\gamma_{tk})_{t,k=1}^T = (1-\beta) \begin{bmatrix}
\eta_1 (1+\beta) & 0                  & \ldots & 0 \\
\eta_1 \beta^2   & \eta_2 (1+\beta)   & \ldots & 0 \\
\eta_1 \beta^3   & \eta_2 \beta^2     & \ldots & 0 \\
\vdots           & \vdots             & \ddots & 0 \\
\eta_1 \beta^T   & \eta_2 \beta^{T-1} & \ldots & \eta_T (1+\beta)
\end{bmatrix}.
\]
Next we verify that Nesterov momentum satisfies the assumptions required for the
general theory, and make the relevant constants explicit.

\begin{lemma}
\label{lem:nesterov}
Nesterov momentum \eqref{eq:nesterov} with nonincreasing positive step sizes
$\eta_1 \geq \eta_2 \geq \cdots$ and $\beta \in [0,1)$ satisfies
\cref{ass:positive-total-weights, ass:bound-stepsize-scheme} with $\delta =
\eta_1 \beta^2\!/(1-\beta)$ and $\rho_t = \eta_t$ for all $t \in \N$.
\end{lemma}

\begin{proof}
Since $\eta_k > 0$ and $\beta \in [0,1)$, the coefficients $\gamma_{tk}$ are
nonnegative. To verify the remaining condition in
\cref{ass:bound-stepsize-scheme}, note that
\begin{align*}
\sum_{k=1}^{t-1} \sum_{u=t}^\infty \gamma_{uk}
&= \sum_{k=1}^{t-1} \eta_k (1-\beta) \sum_{u=t}^\infty \beta^{u-k+1} \\
&= \sum_{k=1}^{t-1} \eta_k \beta^{t-k+1} \\
&\leq \eta_1 \sum_{j=2}^t \beta^j \\
&\leq \eta_1 \sum_{j=2}^{\infty} \beta^j \\
&= \frac{\eta_1\beta^2}{1-\beta},
\end{align*}
i.e., we may take $\delta = \eta_1\beta^2\!/(1-\beta)$. Finally,
\[
\rho_k
  = \eta_k (1-\beta) \biggl(1 + \beta + \sum_{j=1}^\infty \beta^{j+1} \biggr)
  = \eta_k.\qedhere
\]
\end{proof}

Now that we have verified \cref{ass:positive-total-weights,
ass:bound-stepsize-scheme}, with constants quite similar to the heavy ball case
(\cref{lem:hb}), translation of the general theory to Nesterov momentum carries
over just as in \cref{sec:hb}, though we omit exact bounds because they are very
similar to the heavy ball case, only differing in small constant factors.

\section[Proofs for iterate smoothness and related results]
{Proofs for \cref{sec:smoothness} and related results}
\label{app:smoothness}

\subsection{General smoothness result}

\begin{proposition}
\label{prop:momentum-smooth}
Fix $T \geq 3$ and $\beta \in [0,1)$. Let $r_1,\ldots,r_{T-1} \in \R^d$, $z_1
\in \R^d$, and $m_0 = 0$. For $t=1,\ldots,T-1$, let
\begin{align*}
m_t &= \beta m_{t-1}+(1-\beta)r_t, \\
z_{t+1} & = z_t-m_t.
\end{align*}
If \smash{$\max_{t \leq T-1} \norm{r_t}_2 = 0$}, then \smash{$\frac{1}{T-2}
\sum_{t=2}^{T-1} \norm{(\Delta^{\!2} z)_{t-1}}_2 = 0$}. Otherwise,
\begin{align*}
\frac{1}{T-2} \sum_{t=2}^{T-1} \norm{(\Delta^{\!2} z)_{t-1}}_2
&\leq \frac{2(1-\beta)}{1+\beta} \max_{t \leq T-1} \norm{r_t}_2 +
\frac{\norm{z_2-z_1}_2^2-\norm{z_T-z_{T-1}}_2^2}
{2(T-2) \max_{t \leq T-1} \norm{r_t}_2} \\
&\leq \biggl(\frac{2(1-\beta)}{1+\beta} +
\frac{(1-\beta)^2}{2(T-2)}\biggr)\max_{t \leq T-1} \norm{r_t}_2.
\end{align*}
The first inequality is tight in the sense that equality is achieved by at least
one input sequence.
\end{proposition}

\begin{proof}
Let \smash{$B = \max_{t \leq T-1} \norm{r_t}_2$}. If $B=0$, then $r_t=0$ for all
$t=1,\ldots,T-1$, so $m_t=0$ for all $t=0,\ldots,T-1$ and the smoothness penalty
is zero. Assume $B>0$. For $t=2,\ldots,T-1$, let $v_t = r_t-m_{t-1}$. Now since
$(\Delta^{\!2} z)_{t-1} = -(m_t-m_{t-1}) = -(1-\beta)v_t$,
\begin{align*}
2 B \norm{(\Delta^{\!2} z)_{t-1}}_2 +\norm{m_t}_2^2-\norm{m_{t-1}}_2^2
&=2(1-\beta) B \norm{v_t}_2 + 2(1-\beta)m_{t-1}^\T v_t
+(1-\beta)^2\norm{v_t}_2^2 \\
&=2(1-\beta) B \norm{v_t}_2 + 2(1-\beta)r_t^\T v_t
- (1-\beta)(1+\beta)\norm{v_t}_2^2 \\
&\leq 4(1-\beta) B \norm{v_t}_2 - (1-\beta)(1+\beta)\norm{v_t}_2^2 \\
&\leq \frac{4(1-\beta)}{1+\beta} B^2,
\end{align*}
where the second-to-last inequality follows by Cauchy--Schwarz and $\norm{r_t}_2
\leq B$, and the last inequality follows by maximizing over $\norm{v_t}_2 \geq
0$. Averaging over $t=2,\ldots,T-1$ and using
$\norm{m_t}_2=\norm{z_{t+1}-z_t}_2$ gives the first inequality in the
proposition:
\[
\frac{1}{T-2} \sum_{t=2}^{T-1} \norm{(\Delta^{\!2} z)_{t-1}}_2
\leq \frac{2(1-\beta)}{1+\beta} B + \frac{\norm{z_2-z_1}_2^2
-\norm{z_T-z_{T-1}}_2^2}{2(T-2) B}.
\]
The second follows by dropping $-\norm{z_T-z_{T-1}}_2^2 \leq 0$, and using
$\norm{z_2-z_1}_2=(1-\beta)\norm{r_1}_2 \leq (1-\beta) B$.

To verify tightness of the first inequality, fix $p \in \R^d$ with
$\norm{p}_2=1$ and define $r_1 = -p/(1+\beta)$, and $r_t = (-1)^t p$ for
$t=2,\ldots,T-1$. Then $B=1$. We proceed by induction on $t$ to prove
$m_t=(-1)^t(1-\beta)p/(1+\beta)$ for $t=1,\ldots,T-1$. For the base case $t=1$,
we have $m_1=(1-\beta)r_1=-p(1-\beta) / (1+\beta)$. For the inductive step,
assume the inductive hypothesis holds at $t-1$, so that
$m_{t-1}=(-1)^{t-1}(1-\beta)p/(1+\beta)$. Then,
\[
m_t = \beta(-1)^{t-1}\frac{1-\beta}{1+\beta}p+(1-\beta)(-1)^t p
= (-1)^t\frac{1-\beta}{1+\beta}p.
\]
Therefore $\norm{(\Delta^{\!2} z)_{t-1}}_2 = 2(1-\beta)/(1+\beta)$ for each
$t=2,\ldots,T-1$, while $\norm{z_2-z_1}_2 = \norm{z_T-z_{T-1}}_2 =
(1-\beta)/(1+\beta)$, so the first inequality holds with equality.
\end{proof}

\subsection[Proof of the corollary for heavy ball]
{Proof of \cref{cor:hb-smooth}}

Let $r_t = \eta_t g_t(\theta_t)$ for $t=1,\ldots,T-1$. With $m_0 = 0$ and $m_t =
\theta_t-\theta_{t+1} = (1-\beta) \sum_{k=1}^t \beta^{t-k}r_k$, we have
$m_t=\beta m_{t-1}+(1-\beta)r_t$, and the corollary follows by applying
\cref{prop:momentum-smooth}. The tightness claim follows by noticing that the
tight input sequence in \cref{prop:momentum-smooth} is realized by linear losses
$\ell_t(\theta) = (r_t/\eta_t)^\T \theta$.

\subsection{Nesterov momentum}

\begin{corollary}
\label{cor:nesterov-smooth}
Consider Nesterov momentum \eqref{eq:nesterov} with positive step sizes
$(\eta_t)_{t \in \N}$, $\beta \in [0,1)$, and arbitrary initialization $\theta_1
\in \R^d$. Fix any $T \geq 3$. Let $u_0 = 0$ and $u_t = \eta_t g_t(\omega_t)$
for $t=1,\ldots,T-1$. Let $ r_t = (1+\beta)u_t-\beta u_{t-1}$ for
$t=1,\ldots,T-1$. If \smash{$\max_{t \leq T-1} \norm{r_t}_2 = 0$}, then the
smoothness penalty is zero. Otherwise,
\begin{align*}
\frac{1}{T-2}\sum_{t=2}^{T-1} \norm{(\Delta^{\!2} \omega)_{t-1}}_2
&\leq \frac{2(1-\beta)}{1+\beta} \max_{t \leq T-1} \norm{r_t}_2
+ \frac{\norm{\omega_2-\omega_1}_2^2 -\norm{\omega_T-\omega_{T-1}}_2^2}
{2(T-2)\max_{t \leq T-1} \norm{r_t}_2} \\
&\leq \biggl(\frac{2(1-\beta)}{1+\beta} + \frac{(1-\beta)^2}
{2(T-2)}\biggr)\max_{t \leq T-1} \norm{r_t}_2.
\end{align*}
The first inequality is tight in the sense that equality is achieved by at least
one problem instance. Moreover,
\[
\frac{1}{T-2}\sum_{t=2}^{T-1} \norm{(\Delta^{\!2} \omega)_{t-1}}_2
\leq \biggl(\frac{2(1-\beta)(1+2\beta)}{1+\beta} +
\frac{(1-\beta)^2(1+\beta)}{2(T-2)}\biggr)
\max_{t \leq T-1} \eta_t \norm{g_t(\omega_t)}_2.
\]
\end{corollary}

\begin{proof}
Recall that for Nesterov momentum, for all $k \leq t$, we have $\gamma_{tk} =
\eta_k(1-\beta)(\mathbb{I}\{t=k\}+\beta^{t-k+1})$. Therefore, for all
$t=1,\ldots,T-1$,
\[
m_t = (1-\beta)\biggl((1+\beta)u_t +\sum_{k=1}^{t-1}\beta^{t-k+1}u_k\biggr),
\]
where $m_t = \omega_t-\omega_{t+1}$. It follows that, for all $t=1,\ldots,T-1$,
\[
m_t=\beta m_{t-1}+(1-\beta)((1+\beta)u_t-\beta u_{t-1}),
\]
where $m_0=0$ and $u_0=0$. Applying \cref{prop:momentum-smooth} then gives the
first two inequalities. The first is tight, since the tight input sequence in
the proposition is realized by setting $u_0=0$, $u_t=(r_t+\beta
u_{t-1})/(1+\beta)$ for $t=1,\ldots,T-1$, and taking linear losses
$\ell_t(\theta)=(u_t/\eta_t)^\T \theta$.

For the third inequality, let $B = \max_{t \leq T-1} \norm{\eta_t
g_t(\omega_t)}_2$ and $R = \max_{t \leq T-1} \norm{r_t}_2$. If $R = 0$, the
result follows from the first case in the corollary. Otherwise, we drop the term
$-\norm{\omega_T-\omega_{T-1}}_2^2 \leq 0$ from the first inequality and use
$\norm{\omega_2-\omega_1}_2 = (1-\beta)\norm{r_1}_2$, $\norm{r_1}_2^2/R \leq
\norm{r_1}_2 \leq (1+\beta)B$, and
\[
R \leq (1+2\beta)B.
\]
Plugging in these bounds establishes the result.
\end{proof}

\subsection{Comparison to path norm}
\label{app:path-norm}

An alternative to \eqref{eq:smooth-penalty} would be to consider a penalty based
on what is sometimes called the \emph{path norm} in the online learning
literature:
\begin{equation}
\label{eq:path-norm}
\frac{1}{T-1} \sum_{t=1}^{T-1} \norm{(\Delta \theta)_t}_2
  = \frac{1}{T-1} \sum_{t=1}^{T-1} \norm{\theta_{t+1} - \theta_t}_2.
\end{equation}
Here each $(\Delta \theta)_t = \theta_{t+1} - \theta_t$ is the first difference
of the sequence at $t$, which may be seen as the discrete first derivative of
the sequence at $t$. In other words, while \eqref{eq:smooth-penalty} can be seen
as measuring discrete second-order total variation, the penalty in
\eqref{eq:path-norm} measures discrete first-order total variation.

Why study \eqref{eq:smooth-penalty} instead of \eqref{eq:path-norm}? Simply put,
we should expect the iterate sequence
$\theta_t,\theta_{t+1},\theta_{t+2},\ldots$ to change continually from one
iteration to the next, because to remain static would mean we are no longer
learning. Thus, the path norm, which measures variation in the iterates
themselves, seems less appropriate for our purposes compared to the second-order
penalty in \eqref{eq:smooth-penalty}, which measures variation in the rate of
change:
\[
(\Delta^{\!2} \theta)_{t-1} = (\Delta \theta)_t - (\Delta \theta)_{t-1} =
(\theta_{t+1} - \theta_t) - (\theta_t - \theta_{t-1}).
\]
Large values of $\norm{(\Delta^{\!2} \theta)_{t-1}}_2$ precisely correspond to
sharp changes in direction, which is tightly connected to the jaggedness we want
to control.

\subsection{Comparison to post-hoc smoothing}
\label{app:post-hoc}

To achieve smoother iterate paths, a seemingly natural idea would be to apply
some type of post-hoc smoothing to ordinary gradient descent iterates. However,
this can badly disrupt gradient equilibrium guarantees. As a simple example,
consider $\ell_t(\theta) = \psi_\tau(-\theta)$ for $t \in \N$, where $\tau =
1/3$. Given a step size $\eta > 0$ for gradient descent, consider initializing
at $\theta_1 = \varepsilon$ with $0 < \varepsilon < \eta/3$. Then the gradient
descent updates are:
\begin{align*}
\theta_2 &= \theta_1 - \eta (1 - \tau) = \varepsilon - \frac{2\eta}{3}, \\
\theta_3 &= \theta_2 + \eta \tau = \varepsilon - \frac{\eta}{3}, \\
\theta_4 &= \theta_3 + \eta \tau = \varepsilon,
\end{align*}
and the iterates will continue to cycle through $\varepsilon - 2\eta/3,
\varepsilon - \eta/3, \varepsilon$ indefinitely. Note that the gradients cycle
through $2/3, -1/3, -1/3$ indefinitely, so \eqref{eq:geq} is obtained.
Meanwhile, it is easy to see that smoothing the iterates using a trailing
average fails to satisfy \eqref{eq:geq}. Supposing the trailing average uses a
window length $w$, if $w$ is a multiple of 3, then the average will always be
$\varepsilon - \eta/3 < 0$, and the gradient at such a point will always be
$-1/3$. In other words, in this case, the average of the gradients of the
smoothed sequence trivially converges to $-1/3$.

Consider instead taking a weighted trailing average of iterates but then using
this to fuel the gradient descent update. That is, $\theta_t = \beta
\theta_{t-1} + (1-\beta) v_t$, and $v_{t+1} = v_t - \eta_t g_t(\theta_t)$. This
is in fact an equivalent representation of heavy ball, as per Proposition 5 in
\citet{sebbouh2021almost}. It provides another perspective on why heavy ball
produces smoother iterate paths.

\section[Details for the experiments and further results]
{Details for \cref{sec:exp} and further results}
\label{app:exp}

All experiments in the main text and in this section were run on a laptop with
an Intel Core i7-8565U CPU (4 cores, 8 threads) and 16 GB of RAM. The full
pipeline, covering both the debiasing and calibration experiments, completes in
under 10 minutes.

In some practical situations, being marginally unbiased might not provide enough
risk mitigation. Returning to the Shasta Reservoir experiment, water inflow
predictions might be unbiased in the long term even if they are always highly
biased upward during hot periods and highly biased downward in cold periods.
This would clearly be problematic, and thus, it would be desirable to achieve
unbiasedness conditionally on the calendar month of the response. We can achieve
this by simply performing separate online updates in each of the five calendar
months (February through June) in which the response is measured. This is
effectively accomplished by running (in parallel) five separate instances of
\cref{alg:debiasing}, with gradients set to zero outside their respective
months. Equivalently, we may consider this as running heavy ball on the
regression loss $\ell_t(\theta) = \frac12 (z_t - f_t - w_t^\T \theta)^2$, where
$\theta \in \R^5$ and $w_t \in \{0,1\}^5$ is a feature indicating the month of
time $t$. \citet{angelopoulos2025gradient} refer to this as \emph{multigroup}
debiasing, and it is a special case of more general decorrelation
(multiaccuracy-type) problems which they study.

\begin{figure}[t]
\centering
\begin{subfigure}[b]{0.5\textwidth}
\centering
\includegraphics{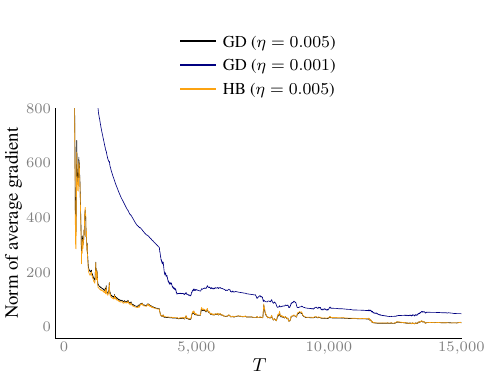}
\caption{Norm of average gradients.}
\label{fig:gradients-squared-loss-multigroup}
\end{subfigure}%
\begin{subfigure}[b]{0.5\textwidth}
\centering
\includegraphics{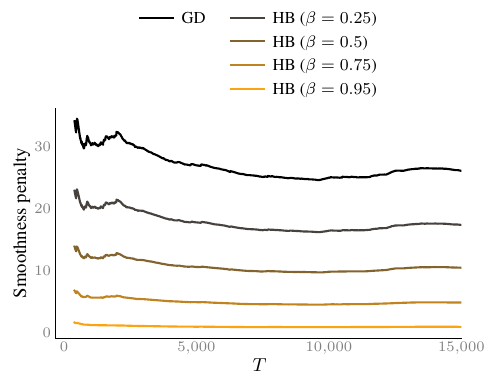}
\caption{Smoothness penalty.}
\label{fig:smoothness-squared-loss-multigroup}
\end{subfigure}
\caption{Multigroup debiasing. HB uses $\beta = 0.95$ in (a), and all methods
  use $\eta = 0.005$ in (b). For larger $\beta$, HB achieves a much lower
  smoothness penalty at comparable gradient equilibrium error.}
\label{fig:multigroup}
\end{figure}

\cref{fig:multigroup} displays the results, which are broadly similar to those
in \cref{fig:gradients-squared-loss, fig:smoothness-squared-loss}: HB is able to
achieve much smoother paths at comparable gradient equilibrium error.